\documentclass[10pt, conference, letterpaper]{IEEEtran}
\IEEEoverridecommandlockouts

\usepackage{epsfig}
\usepackage{epstopdf}
\usepackage{subfigure}
\usepackage{times}
\usepackage{cite}
\usepackage{enumerate}
\usepackage{wrapfig}
\usepackage{algorithm}
\usepackage{algorithmic}
\usepackage[algo2e]{algorithm2e}
\usepackage{balance}
\usepackage{graphicx}
\usepackage{footmisc}
\usepackage{graphicx}
\usepackage{amssymb}
\usepackage{amsbsy}
\usepackage{amsmath}
\usepackage{url}
\usepackage{bm}
\usepackage{bbm}
\usepackage{setspace}

\usepackage{color}
\usepackage{array,cases}
\usepackage{multirow}
\usepackage{theorem}
\usepackage{paralist}
\usepackage{comment}

\usepackage[top = 0.7515in,
bottom = 1.043in,
left = 0.639in,
right = 0.639in]{geometry}

\newcommand{\bp}{\ensuremath{{\bm p}}}

\newcommand{\set}[1]{\ensuremath{\mathcal #1}}

\newcommand{\separator}{
  \begin{center}
    \rule{\columnwidth}{0.3mm}
  \end{center}
}

\theorembodyfont{\rmfamily}
\theoremheaderfont{\bfseries\slshape}
\newtheorem{theorem}{Theorem}

\newtheorem{remark}{\underline{Remark}}

\newtheorem{definition}{Definition}

\newcommand{\prob}[1]{\mathbb{P}[ #1 ]}
\newcommand{\expect}[1]{\mathbb{E}\left[ #1 \right]}

\newcommand{\beq}{\begin{eqnarray*}}
\newcommand{\eeq}{\end{eqnarray*}}
\newcommand{\beqn}{\begin{eqnarray}}
\newcommand{\eeqn}{\end{eqnarray}}
\newcommand{\bemn}{\begin{multiline}}
\newcommand{\eemn}{\end{multiline}}

\newcommand\ie{{\em i.e.}}

\def\N{\mathbb{N}}
\def\R{\mathbb{R}}

\def\ind{{\bf 1}}

\def\PP{{\mathrm P}}
\def\EE{{\mathrm E}}

\def\bp{\noindent{\it Proof.}\ }
\def\ep{\hfill $\Box$}

\newcommand{\ud}{\mathrm{d}}

\def\ind{\mathbbm{1}}
\def\Pcal{\mathcal{P}}

\def\PP{\mathbb{P}}
\def\EE{\mathbb{E}}

\makeatletter
\def\ps@headings{%
\def\@oddhead{\mbox{}\scriptsize\rightmark \hfil \thepage}%
\def\@evenhead{\scriptsize\thepage \hfil \leftmark\mbox{}}%
\def\@oddfoot{}%
\def\@evenfoot{}}

\newcommand{\Rmnum}[1]{\expandafter\@slowromancap\romannumeral #1@}
\makeatother
\newtheorem{proposition}[theorem]{Proposition}
\newtheorem{lemma}[theorem]{Lemma}

\begin{document}
\sloppy

\title{Cluster-Aided Mobility Predictions}

\author{Jaeseong Jeong$^\dag$, Mathieu Leconte$^\ddag$ and Alexandre Proutiere$^\dag$ 
\thanks{
$\dag$: KTH, Royal Institute of Technology, EE School / ACL, Osquldasv. 10, Stockholm 100-44, Sweden, email: \{jaeseong,alepro\}@kth.se. 
$\ddag$: Huawei, France, email: mathieu.leconte@huawei.com 
}}

\maketitle

\begin{abstract}
Predicting the future location of users in wireless networks has numerous applications, and can help service providers to improve the quality of service perceived by their clients. The location predictors proposed so far estimate the next location of a specific user by inspecting the past individual trajectories of this user. As a consequence, when the training data collected for a given user is limited, the resulting prediction is inaccurate. In this paper, we develop {\it cluster-aided} predictors that exploit past trajectories collected from {\it all} users to predict the next location of a given user. These predictors rely on clustering techniques and extract from the training data similarities among the mobility patterns of the various users to improve the prediction accuracy. Specifically, we present CAMP (Cluster-Aided Mobility Predictor), a cluster-aided predictor whose design is based on recent non-parametric bayesian statistical tools. CAMP is robust and adaptive in the sense that it exploits similarities in users' mobility only if such similarities are really present in the training data. We analytically prove the consistency of the predictions provided by CAMP, and investigate its performance using two large-scale datasets. CAMP significantly outperforms existing predictors, and in particular those that only exploit individual past trajectories.

\end{abstract}

\section{Introduction}

Predicting users' mobility in wireless networks has received a great deal of attention recently, strongly motivated by a wide range of applications. Examples of such applications include: location-based services provided to users by anticipating their movements (e.g., mobile advertisement, recommendation systems, risk alarm); urban traffic engineering and forecasting; the design of more efficient radio resource allocation protocols (e.g., scheduling and handover management~\cite{Breadcrumbs08}, data prefetching~\cite{siris2013enhancing} and energy efficient location sensing~\cite{chon2012evaluating}). However, for these applications to significantly benefit from users' mobility predictions, the latter should be made with a sufficiently high degree of accuracy.

Many mobility prediction methods and algorithms have been devised over the last decade, see e.g.  \cite{song2006evaluating,scellato2011nextplace,chon2012evaluating,jacquet2002universal}. The algorithms proposed so far estimate the next location of a specific user by inspecting the data available about her past mobility, i.e., her past trajectory, and exploit the inherent repeated patterns present in this data. These patterns correspond to the regular behavior of the user, e.g. commuting from home to work or visiting favourite restaurants, and need to be extracted from the data to provide accurate predictions. To this aim, one has to observe the behavior of the user over long periods of time. Unfortunately, gathering data about users' mobility can be quite challenging. For instance, detecting the current location of a user with sensors (e.g., GPS, Wi-Fi and cell tower) consumes a non-negligible energy. Users may also hesitate to log their trajectories to preserve their privacy. In any case, when the data about the mobility of a given user is limited, it is hard to identify her typical mobility patterns, and in turn difficult to provide accurate predictions on her next move or location.

In this paper, we aim at devising mobility predictors that perform well even if the past trajectories gathered for the various users are short. Our main idea is to develop {\it cluster-aided} predictors that exploit the data (i.e., past trajectories) collected from {\it all} users to predict the next location of a given user. These predictors rely on clustering techniques and extract from the training data similarities among the mobility patterns of the various users to improve the prediction accuracy. More precisely, we make the following contributions: 

\begin{itemize}
\item We present CAMP (Cluster-Aided Mobility Predictor), a cluster-aided predictor whose design is based on recent non-parametric bayesian statistical tools \cite{ferguson1973bayesian,mcauliffe2006nonparametric}. CAMP extracts, from the data, clusters of users with similar mobility processes, and exploit this clustered structure to provide accurate mobility predictions. The use of non-parametric statistical tools allows us to adapt the number of extracted clusters to the training data (this number can actually grow with the data, i.e., with the number of users). This confers to our algorithm a strong robustness, i.e., CAMP exploits similarities in users' mobility only if such similarities are really present in the training data. 
\item We derive theoretical performance guarantees for the predictions made under the CAMP algorithm. In particular, we show that CAMP can achieve the performance of an optimal predictor (among the set of all predictors) when the number of users grows large, and for a large class of mobility models. 
\item Finally, we compare the performance of our predictor to that of other existing predictors using two large-scale mobility datasets (corresponding to a Wi-Fi and a cellular network, respectively). CAMP significantly outperforms existing predictors, and in particular those that only exploit individual past trajectories to estimate users' next location. 

\end{itemize}

\section{Related work}

Most of existing mobility prediction methods estimate the next location of a specific user by inspecting the past individual trajectories of this user. One of the most popular mobility predictors consists in modelling the user trajectory as an order-$k$ Markov chain. Predictors based on the order-$k$ Markov model are asymptotically optimal \cite{merhav1993some,jacquet2002universal} for a large class of mobility models. This optimality only holds asymptotically when the length of the observed user past trajectory tends to infinity. Unfortunately, when the observed past trajectory of the user is rather short, these predictors perform poorly. Such phenomenon is often referred to as the ``cold-start problem". To improve the performance of these predictors for short histories, a fallback mechanism can be added \cite{song2006evaluating} to reduce the order of the Markov model when the current sequence of $k$ previous locations has not been encountered before. Alternatively, one may adapt the order of the Markov model used for prediction as in the Sampled Pattern Matching (SPM) algorithm \cite{jacquet2002universal}, which sets the order of the Markov model to a fraction of the longest suffix match in the history. SPM is asymptotically optimal with provable bounds on its rate of convergence, when the trajectory is generated by a stationary mixing source. Another type of mobility predictor, Nextplace~\cite{scellato2011nextplace} attempts to leverage the time-stamps that may be associated with the successive locations visited by the user. 
Empirical evaluations \cite{song2006evaluating,chon2012evaluating} show that complex mobility models do not perform well: the order-$2$ Markov predictor with fallback gives comparable performance to that of SPM~\cite{jacquet2002universal}, NextPlace~\cite{scellato2011nextplace} and higher order Markov predictors. In addition \cite{chon2012evaluating} reports that the order-$1$ Markov predictor can actually provide better predictions than higher order Markov predictors, as the latter suffer more from the lack of training data.

There have been a few papers aiming at clustering trajectories or more generally stochastic processes. For example, \cite{smyth1997clustering} proposes algorithms to find clusters of trajectories based on likelihood maximization for an underlying hidden Markov model. For the same problem, \cite{jebara2007spectral} uses spectral clustering in a semi-parametric manner based on Bhattacharyya affinity metric between pairs of trajectories. Those methods would not work well in our setting. This is due to the facts that (i) users belonging to a same cluster should have trajectories generated by identical parameters, and (ii) the number of clusters should be known beforehand, or estimated in a reliable way. The non-parametric Bayesian approach developed in this paper addresses both issues. \cite{mcinerney2013modelling} also introduced Bayesian approach that focused on the similarity between users' temporal patterns. But they do not consider the similarity between spatial trajectories and the correlation to the recent locations which are crucial to the correct predictions in our setting.

\section{Models and Objectives}\label{sec: model}
\label{sec:problem}

In this section, we first describe the data on past user trajectories available at a given time to build predictors. We then provide a model for user mobility, used to define our non-parametric inference approach, as well as its objectives.

\subsection{Collected Data}

We consider the problem of predicting at a given time the mobility, i.e., the next position of users based on observations about past users' trajectories. These observations are collected and stored on a server. The set of users is denoted by $\set U$, and users are all moving within a common finite set $\set L$ of $L$ locations. The trajectory collected for user $u$ is denoted by $x^u=(x_1^u,\ldots, x_{n^u}^u)$, where $x_t^u$ corresponds to the $t$-th location visited by user $u$, and where $n^u$ refers to the length of the trajectory. $x_{n^u}^u$ denotes the current location of user $u$. By definition, we impose $x_t^u\neq x_{t+1}^u$, i.e., two consecutive locations on a trajectory must be different. Let $x^{\cal U}=(x^u)_{u\in {\cal U}}$ denote the set of user trajectories. Observe that the lengths of the trajectories may vary across users.  
If the location of a user is sensed periodically, we can collect the time a given user has stayed at each location. Those staying times for user $u$ are denoted by $s^u = (s_1^u,\ldots,s_{n^u-1}^u)$, where $s_t^u$ is the staying time at the $t$-th visited location. To simplify the presentation, we present our prediction methods ignoring the staying times $s^u$; but we mention how to extend our approach to include staying times in \textsection\ref{sec:staying}.

Next we introduce additional notations. We denote by $n^u_{i,j}$ the number of observed transitions for user $u$ from location $i$ to $j$, (i.e., $n^u_{i,j}=\sum_{t=1}^{n^u-1}\ind(x^u_t=i,\:x^u_{t+1}=j)$). Similarly, $n^u_i=\sum_{t=1}^{n^u}\ind(x^u_t=i)$ is the number of times user $u$ has been observed at location $i$. Let $\set H\subseteq\cup_{n=0}^\infty\set L^n$ denote the set of all possible trajectories of a given user, and let $\set H^\set U$ be the set of all possible set of trajectories of users in  $\set U$.

\subsection{Mobility Models}

The design of our predictors is based on a simple mobility model. We assume that user trajectories are order-1 Markov chains, with arbitrary initial state or location. More precisely, user-$u$'s trajectory is generated by the transition kernel $\theta^u=(\theta_{i,j}^u)_{i, j\in {\cal L}}\in [0,1]^{L\times L}$, where $\theta_{i,j}^u$ denotes the probability that user $u$ moves from location $i$ to $j$ along her trajectory. Hence, given her initial position $x_1^u$, the probability of observing trajectory $x^u$ is $P_{\theta^u}(x^u):=\prod_{t=1}^{n^u-1}\theta^u_{x^u_t,x^u_{t+1}}$. Our mobility model can be readily extended to order-$k$ Markov chains. However, as observed in \cite{chon2012evaluating}, order-1 Markov chain model already provides reasonably accurate predictions in practice, and higher-order models would require a fall-back mechanism\footnote{To accurately predict the next position of user $u$ given that the sequence of her past $k$ positions is $i_1,\ldots,i_k$, her trajectory should contain numerous instances of this sequence, which typically does not occur if the observed trajectory is short -- and this is precisely the case we are interested in.}\cite{song2006evaluating}. Throughout the paper, we use uppercase letters to represent random variables and the corresponding lowercase letters for their realizations, e.g. $X^u$ (resp. $x^u$) denotes the random (resp. realization of) trajectory of user $u$. 



\subsection{Bayesian Framework, Clusters,
and	 Objectives}\label{subsec:bayes}

We adopt a Bayesian framework, and assume that the transition kernels of the various users are drawn independently from the same distribution $\mu\in\Pcal(\Theta)$\footnote{${\cal P}(\cal M)$ denotes the set of distributions over the set ${\cal M}$, and $\Theta=\{\theta\in [0,1]^{L\times L}: \forall i, \sum_j\theta_{ij}=1\}$.} referred to as the {\em prior} distribution over the set of all possible transition kernels $\Theta$. This assumption is justified by De Finetti's theorem (see \cite{kallenberg2002foundations}, Theorem~11.10) if $(\theta^u)_{u\in {\cal U}}$ are exchangeable (which is typically the case if users are a priori indistinguishable). In the following, the expectation and probability under $\mu$ are denoted by $\EE$ and $\PP$, respectively. To summarize, the trajectories of users are generated using the following hierarchical model: for all $u\in {\cal U}$, $\theta_u\sim \mu$, $X^u \sim P_{\theta^u}$, and $n^u, X_1^u$ are arbitrarily fixed.

To provide accurate predictions even if observed trajectories are rather short, we leverage similarities among user mobility patterns. It seems reasonable to think that the trajectories of some users are generated through similar transition kernels. In other words, the distribution $\mu$ might exhibit a clustered structure, putting mass around a few typical transition kernels. Our predictors will identify these clusters, and exploit this structure, i.e., to predict the next location of a user $u$, we shall leverage the observed trajectories of all users who belong to user-$u$'s cluster.  

For any user $u$, we aim at proposing an accurate predictor $\hat{x}^u\in {\cal L}$ of her next location, given the observed trajectories $X^{\cal U}=x^{\cal U}$ of all users. The (Bayesian) accuracy of a predictor $\hat x^u$ for user $u$, denoted by $\pi^u(\hat x^u)$, is defined as $ \pi^u(\hat x^u) := \PP\left(X^u_{n^u+1}=\hat x^u|x^\set U\right)=\EE[\theta^u_{x^u_{n^u},\hat x^u}|x^\set U]$ (where for conciseness, we write $\PP\left(\cdot | x^\set U\right)=\PP\left(\cdot | X^\set U = x^\set U\right)$). Clearly, given $X^\set U = x^\set U$, the best possible predictor would be:
\begin{equation}
\hat x^u \in \arg \max_{j\in\set L} \EE[\theta^u_{x^u_{n^u},j}|x^{\cal U}].\label{eq:ftopt}
\end{equation}
Computing this optimal predictor, referred to as the Bayesian predictor with prior $\mu$, requires the knowledge of $\mu$. Indeed:
\begin{equation}\label{eq:opt2}
\EE[\theta^u_{i,j}|x^{\cal U}]=\frac{\int_\theta \theta_{i,j} P_\theta(x^u)\mu(\ud\theta)}{\int_\theta P_\theta(x^u)\mu(\ud\theta)}.
\end{equation}
Since here the prior distribution $\mu$ is unknown, we will first estimate $\mu$ from the data, and then construct our predictor according to \eqref{eq:ftopt}-\eqref{eq:opt2}.

\section{Bayesian Non-parametric Inference}\label{sec: algo}

In view of the model described in the previous section, we can devise an accurate mobility predictor if we are able to provide a good approximation of the prior distribution $\mu$ on the transition kernels dictating the mobility of the various users. If $\mu$ concentrates its mass around a few typical kernels that would in turn define clusters of users (i.e., users with similar mobility patterns), we would like to devise an inference method identifying these clusters. On the other hand, our inference method should not discover clusters if there are none, nor specify in advance the number of clusters (as in the traditional mixture modelling approach). Towards these objectives, we apply a Bayesian non-parametric approach that estimates how many clusters are needed to model the observed data and also allows the number of clusters to grow with the size of the data. In Bayesian non-parametric approaches, the complexity of the model (here the number of clusters) is part of the posterior distribution, and is allowed to grow with the data, which confers flexibility and robustness to these approaches.     
In the remaining of this section, we first present an overview of the Dirichlet Process mixture model, a particular Bayesian non-parametric model, and then apply this model to the design of CAMP (Cluster-Aided Mobility Predictor), a robust and flexible prediction algorithm that efficiently exploits similarities in users' mobility, if any exist.  

\subsection{Dirichlet Process Mixture Model}

When applying Bayesian non-parametric inference techniques \cite{ferguson1973bayesian} to our prediction problem, we add one level of randomness. More precisely, we approximate the prior distribution $\mu$ on the transition kernels $\theta^u$ by a random variable $\hat\mu$ with distribution $g\in {\cal P}({\cal P}(\Theta))$. This additional level of randomness allows us to introduce some flexibility in the number of clusters present in $\mu$. We shall compute the posterior distribution $g$ given the observations $x^{\cal U}$, and hope that this posterior distribution, denoted as $g|x^{\cal U}$, will concentrate its mass around the true prior distribution $\mu$. To evaluate $g|x^{\cal U}$, we use Gibbs sampling techniques (see Section \ref{subsec: MCMC}), and from these samples, we shall estimate the true prior $\mu$, and derive our predictor by replacing $\mu$ by its estimate in (\ref{eq:ftopt})-(\ref{eq:opt2}). 

For the higher-level distribution $g$, we use the Dirichlet Process (DP) mixture model, a standard choice of prior over infinite dimensional spaces, such as ${\cal P}(\Theta)$. The DP mixture model has a possibly infinite number of mixture components or clusters, and is defined by a {\it concentration} parameter $\alpha>0$, which impacts the number of clusters, and a {\it base} distribution $G_0\in {\cal P}(\Theta)$, from which new clusters are drawn. The DP mixture model with parameters $\alpha$ and $G_0$ is denoted by $DP(\alpha, G_0)$ and defined as follows. If $\nu$ is a random measure drawn from $DP(\alpha,G_0)$ (i.e., $\nu\sim DP(\alpha,G_0)$), and $\{A_1, A_2, \cdots, A_K\}$ is a (measurable) partition of $\Theta$, then $(\nu(A_1),\cdots,\nu(A_K))$ follows a Dirichlet distribution with parameters $(\alpha G_0(A_1),\cdots,\alpha G_0(A_K))$\footnote{The Dirichlet distribution with parameters $(\alpha_1,\ldots,\alpha_K))$ has density (with respect to Lebesgue measure) proportional to $\ind(x_1>0,\ldots,x_K>0)\ind(x_1+\ldots+x_K=1)\prod_{k=1}^Kx_k^{\alpha_k}$.}. It is well known \cite{blackwell1973ferguson} that a sample $\nu$ from $DP(\alpha,G_0)$ has the form $\nu=\sum_{c=1}^\infty\beta^c\delta_{\overline\theta^c},$
where $\delta_\theta$ is the Dirac measure at point $\theta\in\Theta$, the $\overline\theta^c$'s are i.i.d. with distribution $G_0$ and represent the centres of the clusters (indexed by $c$), and the weights $\beta^c$'s are generated using a Beta distribution according to the following stick-breaking construction:
\begin{eqnarray*}
\widetilde\beta^c&\sim&\operatorname{Beta}(1,\alpha)\ (\text{the }\widetilde\beta^c\text{'s are independent}),\\
\beta^c&=&\widetilde\beta^c\prod_{i=1}^{c-1}(1-\widetilde\beta^i).
\end{eqnarray*}
When $(\theta^u)_{u\in {\cal U}}$ is generated under the above DP mixture model, we can compute the distribution of $\theta^u$ given $\theta^{\set U\setminus u}=(\theta^v)_{v\in {\cal U}\setminus\{ u\}}$. When $\theta^{\set U\setminus u}$ is fixed, then users in ${\cal U}\setminus\{ u\}$ are clustered and the set of corresponding clusters is denoted by $c^{{\cal U}\setminus\{ u\}}$. Users in cluster $c\in c^{{\cal U}\setminus\{ u\}}$ share the same transition kernel $\overline\theta^c$, and the number of users assigned to cluster $c$ is denoted by $n_{c,-u}=\sum_{u\in {\cal U}\setminus\{ u\}}\ind_{u\in c}$. The distribution of $\theta^u$ given $\theta^{\set U\setminus u}$ is then:
\begin{eqnarray}
\theta^u|\theta^{\set U\setminus u} &\sim&\left\{\begin{array}{ll}
G_0 & \text{w.p. }\frac{\alpha}{\alpha+|\set U|-1},\\
\delta_{\overline\theta^c} & \text{w.p. }\frac{n_{c,-u}}{\alpha+|\set U|-1}, \forall c\in c^{{\cal U}\setminus\{ u\}}.
\end{array}\right.\label{eqn: clustered property 2}
\end{eqnarray}
(\ref{eqn: clustered property 2}) makes the cluster structure of the DP mixture model explicit. Indeed, when considering a new user $u$, a new cluster containing user $u$ only is created with probability $\frac{\alpha}{\alpha+|\set U|-1}$, and user $u$ is associated with an existing cluster $c$ with probability proportional to the number of users already assigned to this cluster. Refer to \cite{gershman2012tutorial} for a more detailed description on DP mixture models.  

Our prediction method simply consists in approximating $\mathbb{E}[\theta^u|x^{\cal U}]$ by the expectation w.r.t. the posterior distribution $g|x^\set U$. In other words, for user $u$, the estimated next position will be:
\begin{equation}
\hat x^u \in \arg \max_{j\in\set L} E_g[\theta^u_{x^u_{n^u},j}|x^{\cal U}],\label{eq:ftoptTER}
\end{equation}
where $E_g[\cdot]$ denotes the expectation w.r.t. the probability measure induced by $g$. To compute $E_g[\theta^u |x^{\cal U}]$, we rely on Gibbs sampling techniques to generate samples with distribution $g|x^{\cal U}$. The way $g|x^{\cal U}$ concentrates its mass around the true prior $\mu$ will depend on the choice of parameters $\alpha$ and $G_0$, and to improve the accuracy of our predictor, these parameters will be constantly updated when successive samples are produced.


\subsection{CAMP: Cluster-Aided Mobility Predictor}

Next we present CAMP, our mobility prediction algorithm. The objective of this algorithm is to estimate $E_g[\theta^u|x^{\cal U}]$ from which we derive the predictions according to (\ref{eq:ftoptTER}). CAMP consists in generating independent samples of the assignment of users to clusters induced by the posterior distribution $g| x^{\cal U}$, and then in providing an estimate of $E_g[\theta^u|x^{\cal U}]$ from these samples. As mentioned above, the accuracy of this estimate strongly depends on the choice of parameters $\alpha$ and $G_0$ in the DP mixture model, and these parameters will be updated as new samples are generated.

More precisely, the CAMP algorithm consists in two steps. (i) In the first step, we use Gibbs sampler to generate $B$ samples of the assignment of users to clusters under the probability measure induced by $g| x^{\cal U}$, and update the parameters $\alpha$ and $G_0$ of the DP mixture model using these samples (hence we update the prior distribution $g$). We repeat this procedure $K-1$ times. In the $k$-th iteration, we construct $B$ samples of users' assignment. The $b$-th assignment sample is referred to as $c^{{\cal U},b,k}=(c^{u,b,k})_{u\in {\cal U}}$ in CAMP pseudo-code, where $c^{u,b,k}$ is the cluster of user $u$ in that sample. The subroutines providing the assignment samples, and updating the parameters of the prior distribution $g$ are described in details in \S\ref{subsec: MCMC} and \S\ref{subsec: G0 and alpha}, respectively. At the end of the first step, we have constructed a prior distribution $g$ parametrized by $G_0^K$ and $\alpha_K$ which is adapted to the data, i.e., a distribution that concentrates its mass on the true prior $\mu$. (ii) In the second step, we use the updated prior $g$ to generate one last time $B$ samples of users' assignment. Using these samples, we compute an estimate $\hat\theta^u$ of $E_g[\theta^u|x^{\cal U}]$ for each user $u$, and finally derive the prediction $\hat{x}^u$ of the next position of user $u$. The way we compute $\hat\theta^u$ is detailed in \S\ref{subsec: computetheta}.

The CAMP algorithm takes as inputs the data $x^{\cal U}$, the number $K$ of updates of the prior distribution $g$, the number of samples $B$ generated by the Gibbs sampler in each iteration, and the number of times $M$  the users' assignment is updated when producing a single assignment sample using Gibbs sampler (under Gibbs sampler, the assignment is a Markov chain, which we simulate long enough so as it has the desired distribution). $K$, $B$, and $M$ have to be chosen as large as possible. Of course, increasing these parameters also increases the complexity of the algorithm, and we may wish to select the parameters so as to achieve an appropriate trade-off between accuracy and complexity.     

\begin{algorithm}[]
\caption{CAMP \label{algo:CAHB}}
{\bf Input:} $x^\set U, K, B, M$ \\
\underline{Step 1: Updates of $G_0$ and $\alpha$} \\
$G^1_0 \leftarrow \text{Uniform}(\Theta), \alpha_1 \leftarrow 1$  \\
\For{ $k = 1 \ldots K-1$}{
\For{$b = 1 \ldots B$ }{
$c^{\set U,b,k}$ $\leftarrow$ GibbsSampler($x^\set U,G_0^k,\alpha_k,M$) 
}
$G_0^{k+1},\alpha_{k+1}\leftarrow$UpdateDP($ x^{\set U},G_0^k,\{c^{\set U,b,k}\}_{b = 1\ldots B}$)
} 

\underline{Step 2: Last sampling and prediction}\\
\For{$b = 1 \ldots B$ }{
$c^{\set U,b,K}$ $\leftarrow$ GibbsSampler($x^\set U,G_0^K,\alpha_K,M$)
}
Compute $\hat\theta^{u}$ by implementing \eqref{eq:theta_com} using $\{c^{u,b,K}\}_{b=1,\ldots,B}$ and $G_0^K$ \\
$\hat x^u = \arg\max_j \hat\theta^{u}_{x^u_{n^u},j}$

{\bf Output:} $\hat\theta^{u},\hat x^u$
\end{algorithm}

\subsubsection{Sampling from the DP mixture posterior}\label{subsec: MCMC}

We use Gibbs sampler \cite{neal2000markov} to generate independent samples of the assignment of users to clusters under the probability measure induced by the posterior $g| x^{\cal U}$, i.e., samples of assignment with distribution $P_g[c^{\cal U}|x^{\cal U}]$, where $P_g$ denotes the probability measure induced by $g$. Gibbs sampling is a classical MCMC method to generate samples from a given distribution. It consists in constructing and simulating a Markov chain whose stationary state has  the desired distribution. In our case, the state of the Markov chain is the assignment $c^{\cal U}$, and its stationary distribution is $P_g[c^{\cal U}|x^{\cal U}]$. The Markov chain should be simulated long enough (here the number of steps is denoted by $M$) so that at the end of the simulation, the state of the Markov chain has converged to the steady-state. The pseudo-code of the proposed Gibbs sampler is provided in Algorithm \ref{algo:MCMC}, and easily follows from the description of the DP mixture model provided in (\ref{eqn: clustered property 2}).     

To produce a sample of the assignment of users to clusters, we proceed as follows. Initially, we group all users in the same cluster $c_1$, the number of cluster $N$ is set to 1, and the number of users (except for user $u$) $n_{c_1,-u}$ assigned to cluster $c_1$ is $|{\cal U}|-1$. (see Algorithm \ref{algo:MCMC}). Then the assignment is revised $M$ times. In each iteration, each user is considered and assigned to either an existing cluster, or to a newly created cluster (the latter is denoted by $c_{N+1}$ if in the previous iteration there was $N$ clusters). This assignment is made randomly according to the model described in (\ref{eqn: clustered property 2}). Note that in the definition of $\beta_c$, we have $G_0(\ud\theta|x^c)={P_\theta(x^c)G_0( \ud\theta)\over \int_{\theta} P_\theta(x^c)G_0( \ud\theta)}$, where $x^c$ corresponds to the data of users in cluster $c$, i.e., $x^c=(x^u)_{u\in c}$.

\begin{algorithm}[]
\caption{GibbsSampler \label{algo:MCMC}}
{\bf Input:} $x^\set U, G_0, \alpha, M$ \\
$\forall u\in {\cal U}$, $c^u \leftarrow c_1$, $n_{c_1,-u}\leftarrow |{\cal U}|-1$; $N\leftarrow 1$; $c^{\cal U}=\{ c_1 \}$.\\
\For{$i = 1 \ldots M$}{
\For{each $u \in \set{U}$ }{
$c^u \leftarrow c^u\setminus \{ u\}$\\
$\beta_{new}\leftarrow z\frac{\alpha}{\alpha+|\set U|-1}\int_{\theta}P_\theta(x^u)G_0(\ud\theta)$\\
$\beta_c \leftarrow z\frac{n_{c,-u}}{\alpha+|\set U|-1}\int_\theta P_\theta(x^u)G_0(\ud\theta|x^c)$, $\forall c\in c^{{\cal U}\setminus\{ u\}}$ \\
In the above expressions, $z$ is a normalizing constant, i.e., selected so as $\beta_{new}+\sum_{c\in c^{{\cal U}\setminus\{ u\}}}\beta_c=1$; \\
With probability $\beta_{new}$ do:\\
$c_{N+1}\leftarrow \{ u\}$; $c^u\leftarrow c_{N+1}$; $n_{c_{N+1},-u}\leftarrow 0$; $n_{c_{N+1},-v}\leftarrow 1$,  $\forall v\neq u$; $c^{\cal U}\leftarrow c^{\cal U}\cup \{ c_{N+1}\}$; $N\leftarrow N+1$;\\
and with probability $\beta_c$ do:\\
$c^u\leftarrow c$; $c\leftarrow c\cup\{ u\}$; $n_{c,-v} \leftarrow n_{c,-v}+1$, $\forall v\neq u$.
}
}
{\bf Output:} $c^\set U$
\end{algorithm}

\subsubsection{Updates of $G_0$ and $\alpha$}\label{subsec: G0 and alpha}

As in any Bayesian inference method, our prediction method could suffer from a bad choice of parameters $\alpha$ and $G_0$ defining the prior $g$. For example, by choosing a small value for $\alpha$, we tend to get a very small number of clusters, and possibly only one cluster. On the contrary, selecting a too large $\alpha$ would result in a too large number of clusters, and in turn, would make our algorithm unable to capture similarities in the mobility patterns of the various users. To circumvent this issue, we update and fit the parameters to the data, as suggested in \cite{mcauliffe2006nonparametric}. In the CAMP algorithm, the initial base distribution is uniform over all transition kernels (over $\Theta$) and $\alpha$ is taken equal to 1. Then after each iteration, we exploit the samples of assignments of users to clusters to update these initial parameters, by refining our estimates of $G_0$ and $\alpha$. 

\begin{algorithm}[]
\caption{UpdateDP at the $k$-th iteration \label{algo:update}}
{\bf Input:} $ x^\set U,G_0^k, \{c^{\set U,b,k}\}_{b = 1, \ldots, B}$ \\
Compute $G_0^{k+1}(.)$ and $\alpha_{k+1}$ as follows.
\begin{eqnarray}
G_0^{k+1}(.) = \frac{1}{B}\sum_{b=1}^{B} \sum_{c\in c^{\set U,b,k}} \frac{n_{c,b,k}}{|\set{U}|} G_0^k(.|x^c) \label{eq:update G_0}\\
\alpha_{k+1} = \arg \min_{\alpha \in \R} \Big| \sum_{i=1}^{|\set{U}|}\frac{\alpha}{\alpha+i-1} - \frac{1}{B}\sum_{b=1}^{B} N_b\Big | \label{eq:update alpha}
\end{eqnarray}
where $n_{c,b,k}$ is the size of cluster $c\in c^{\set U,b,k}$, and $N_b$ is the total number of (non-empty) clusters in $c^{\set U,b,k}$. \\
{\bf Output:} $G_0^{k+1}, \alpha_{k+1}$
\end{algorithm}

%
Note that (\ref{eq:update G_0}) simply corresponds to a kernel density estimator based on the $B$ cluster samples obtained with prior distribution parametrized by $G_0^k$ and $\alpha_k$, whereas (\ref{eq:update alpha}) corresponds to a maximum likelihood estimate (see \cite{liu1996nonparametric}), which sets $\alpha_{k+1}$ to the value which is most likely to have resulted in the average number of clusters obtained when sampling from the model with parameters $G_0^k$ and $\alpha_k$.

\subsubsection{Computation of $\hat \theta^u$} \label{subsec: computetheta}
 
As mentioned earlier, $\hat\theta^u$ is an estimator of $E_g[\theta^u|x^\set U],$
where $g$ is parameterized by $G_0^K$ and $\alpha_K$, and is used for our prediction of user-$u$'s mobility. $\hat\theta^u$ is just the empirical average of $\bar\theta^c$ for clusters $c$ to which user-$u$ is associated in the $B$ last samples generated in CAMP, i.e.,
\begin{eqnarray}
\hat \theta^u &=& \frac{1}{B}\sum_{b=1}^B E_g[\bar\theta^{c^{u,b,K}}|x^{c^{u,b,K}}] \\
&=& \frac{1}{B}\sum_{b=1}^B \frac{\int_\theta \theta \cdot P_\theta(x^{c^{u,b,K}})G_0^K(d\theta)}{\int_\theta P_\theta(x^{c^{u,b,K}})G_0^K(d\theta)}. \label{eq:theta_com}
\end{eqnarray}
Note that in view of the law of large numbers, when $B$ grows large, $\hat\theta^u$ converges to $E_g[\theta^u|x^\set U]$. The predictions for user $u$ are made by first computing an estimated transition kernel $\hat\theta^u$ according to \eqref{eq:theta_com}. 
We derive an explicit expression of $\hat\theta^u$ that does not depend on $G_0^K$, but only on data and the samples generated in the CAMP algorithms. This expression, given in the following lemma, will be useful to understand to what extent the prediction of user-$u$'s mobility under CAMP leverages observed trajectories of other users.

\begin{lemma} \label{lem: computation}
For any $i,j,$ $\hat \theta^u_{i,j}$ is computed by a weighted sum of all users' empirical transition kernels ($n^v_{i,j}/n^v_i, v \in \set U$), i.e., 
\begin{eqnarray}
\hat \theta^u_{i,j} &=&  \eta_i + \sum_{v\in\set U} \gamma^v_i \frac{n^v_{i,j}}{n^v_i}, \label{eq: hat theta resem} \\
\text{where }\eta_i &=& \sum\limits_{\substack{c_1..c_K:\\u\in c_K}}\xi_{c_1..c_K}\frac{1}{|\set L|+\sum_{k=1}^K n_i^{c_k}}\frac{|\set U|}{n_{c_K}^K}\prod\limits_{k=1}^K\omega_{c_k}^k,  \cr
\gamma^v_i &=& \sum\limits_{\substack{c_1..c_K:\\u\in c_K}}\xi_{c_1..c_K}\frac{n_{i}^{v}\sum_{k=1}^K\ind(v \in c_k)}{|\set L|+\sum_{k=1}^K n_i^{c_k}} \frac{|\set U|}{n_{c_K}^K}\prod\limits_{k=1}^K\omega_{c_k}^k. \nonumber
\end{eqnarray} 
The sum $\sum\limits_{c_1..c_K}$ stands for $\sum_{c_1 \in \set C_1}\cdots \sum_{c_K \in \set C_K}$, and $\set C_k$ is the set of every cluster sampled at $k$-th iterations (i.e., $\set C_k = \{ c | \sum_{b=1}^B \sum_{u \in \set U} \ind(c^{u,b,k}=c) > 0 \}$). $\omega_c^k$ and $\xi_{c_1..c_K}$ are given by:
$$
\xi_{c_1..c_K}=\prod_{i\in\set L}\frac{\prod_{j\in\set L}\Gamma(1+\sum_{k=1..K}n_{i,j}^{c_k})}{\Gamma(|\set L|+\sum_{k=1..K}n_i^{c_k})}, 
$$
$$
\omega_c^{K}=\frac{n_c^{K}}{B|\set U|  \sum\limits_{c_1..c_{K-1}}\xi_{c_1..c_{K-1},c} \prod\limits_{k=1}^{K-1}\omega_{c_k}^k } ,
$$
where $n^c_{i,j}=\sum_{u\in c}n^u_{i,j}$, $n^c_i=\sum_{j\in\set L}n^c_{i,j}$, and $n_c^k=\sum_{b=1}^B\sum_{u\in\set U}\ind(c^{u,b,k}=c)$.  
\end{lemma}
\bp Refer to Appendix. \ep

When the current location $i$ is fixed, the first term in the r.h.s. of \eqref{eq: hat theta resem} is constant over all users. The second term can be interpreted as a weighted sum of the empirical transition kernels of all users (i.e., $n_{i,j}^{v}/n_i^v, \forall v \in \set U$). The weight of user $v$ ($\gamma^v_i$ in \eqref{eq: hat theta resem}) quantifies how much we account for user-$v$'s trajectory in the prediction for user $u$ at the current location $i$, and can be seen as a notion of similarity between $v$ and $u$. Indeed, as the number of sampled clusters in which both $u$ and $v$ are involved increases, $\gamma^v_i$ in \eqref{eq: hat theta resem} increases accordingly. Also, if $v$ has relatively high $n^v_i$ compared to other users (i.e., $v$ has accumulated more observations at the location $i$ than other users), a higher weight is assigned to $v$.

\subsubsection{Estimating the Staying-times}\label{sec:staying}

Next we provide a way of estimating how long user $u$ will stay at her current location $i$. We may perform such estimation when the available data include the time users stay at the various locations.  Typically, the existing spatio-temporal predictors predict the staying time at the current location $x^u_{n^u}$  by computing average~\cite{scellato2011nextplace} or $p$-quantile~\cite{chon2012evaluating} of user $u$'s staying times observed at her previous visits to $x^u_{n^u}.$ On the other hand, CAMP additionally exploits other users' staying time observations using the weight $\gamma^v_i$. More precisely, the staying time of user $u$ at location $x^u_{n^u}$ (denoted by $\hat s^u_{n^u}$) is estimated by 
\begin{align}
\hat s^u_{n^u} =  \sum_{v\in \set U} z\gamma^v_i \frac{1}{n^v_i} \sum_{t:x^v_t = i} s^v_t, \text{ where } i = x^u_{n^u}.
\label{eq: hat_stay}
\end{align}
$z$ in \eqref{eq: hat_stay} is a normalization constant to make the sum of weights over all users equal to 1. The estimate \eqref{eq: hat_stay} is a heuristic, for $\gamma^v_i$ is actually obtained by clustering based on their location trajectories $x^{\set U}$, rather than their staying times. This heuristic estimate actually performs well as empirically shown in Section~\ref{sec: error_staying_time}.

\section{Consistency of CAMP Predictor} \label{sec: Bayesian posterior consistency}

In this section, we analyze to what extent $E_g[\theta^u|x^\set U]$ (that is well approximated, when $B$ is large, by $\hat\theta^u$ derived in the CAMP algorithm) is close to $\EE[\theta^u|x^u]$, the expectation under the true prior $\mu$. We are mainly interested in the regime where the user population $\set U$ becomes large, while the number of observations $n^u$ for each user remains bounded. This regime is motivated by the fact it is often impractical to gather long trajectories for a given user, while the user population available may on the contrary be very large. For the sake of the analysis, we assume that the length $n^u$ of user-$u$'s observed trajectory is a random variable with distribution $p\in {\cal P}(\mathbb{N})$, and that the lengths of trajectories are independent across users. We further assume that the length is upper bounded by $\overline{n}$, e.g., $\overline{n}=\max\{n: p(n)>0\}<\infty$. 

Since the length of trajectories is bounded, we cannot ensure that $|E_g[\theta^u|x^\set U]-\EE[\theta^u|x^u]|$ is arbitrarily small. Indeed, for example if users' trajectories are of length 2 only, we cannot group users into clusters, and in turn, we can only get a precise estimate of the transition kernels averaged over all users. In particular, we cannot hope to estimate $\EE[\theta^u|x^{\cal U}]$ for each user $u$. Next we formalize this observation. We denote by $\set H_{\overline{n}}\subseteq\set L^{\overline{n}}$ the set of possible trajectories of length less than $\overline{n}$. With finite-length observed trajectories, there are distributions $\nu\in {\cal P}(\Theta)$ that cannot be distinguished from the true prior $\mu$ by just observing users' trajectories, i.e., these distributions induce the same law on the observed trajectories as $\mu$: $P_\nu = \mathbb{P}$ on $\set H_{\overline{n}}$ (here $P_\nu$ denotes the probability measure induced under $\nu$, and recall that $\mathbb{P}$ is the probability measure induced by $\mu$). 
We prove that, when the number of observed users grows large, 
$|E_g[\theta^u|x^\set U]-\EE[\theta^u|x^u]|$ is upper-bounded by the performance provided by a distribution $\nu$ indistinguishable from $\mu$, which expresses the consistency of our inference framework.
Before we state our result, we introduce the following two notions:\\
{\it KL $\epsilon$-neighborhood:} the Kullback-Leibler $\epsilon$-neighborhood $K_{\epsilon,\overline{n}}(\mu)$ of a distribution $\mu\in {\cal P}(\Theta)$ with respect to ${\cal H}_{\overline{n}}$ is defined as the following set of distributions:
\begin{equation*}
K_{\epsilon,\overline{n}}(\mu)=\left\{\nu\in\Pcal(\Theta):\:KL_{\overline{n}}(\mu,\nu)<\epsilon\right\},
\end{equation*}
where $KL_{\overline{n}}(\mu,\nu)=\sum_{x\in\set H_{\overline{n}}}P_\mu(x)\log\frac{P_\mu(x)}{P_\nu(x)}$.\\
{\it KL support:} The distribution $\mu$ is in the Kullback-Leibler support of a distribution $g\in\Pcal(\Pcal(\Theta))$ with respect to $\set H_{\overline{n}}$ if  $g(K_{\epsilon,\overline{n}}(\mu))>0$ for all $\epsilon>0$.\\

\begin{theorem}\label{thm: consistency}
If $\mu\in\Pcal(\Theta)$ is in the KL-support of $g$ with respect to $\set H_{\overline{n}}$, then we have, $\mu$-almost surely, for any $i,j\in\set L$,
\begin{equation}
\lim\limits_{|\set U|\to\infty}\left|E_g[\theta^u_{i,j}|X^\set U]-\EE[\theta^u_{i,j}|X^u]\right| \hspace{1.5cm}$$ $$\hspace{1.5cm} \leq\sup\limits_{\substack{\nu\in\Pcal(\Theta)\\P_\nu=\PP\text{ on }\set H_{\overline{n}}}}\left|E_\nu[\theta^u_{i,j}|X^u]-\EE[\theta^u_{i,j}|X^u]\right|. 
\label{eq: consistency}
\end{equation}
\end{theorem}
\bp Refer to Appendix. \ep

The r.h.s. of \eqref{eq: consistency} captures the performance of an algorithm that would perfectly estimate $E_\nu[\theta^u|X^u]$ for the worst distribution $\nu,$ which agrees with the true prior $\mu$ on $\set H_{\overline{n}}.$
Note that in our framework, for the prior $g\in\Pcal(\Pcal(\Theta))$, we use is a DP mixture $DP(G_0,\alpha)$, with a base measure $G_0\in\Pcal(\Theta)$ having full support $\Theta$. Therefore, the KL-support of $g$ is here the whole space $\Pcal(\Theta)$; it thus contains $\mu$.

As far as we are aware, Theorem \ref{thm: consistency} presents the first performance result on inference algorithms using DP mixture models with {\it indirect} observations. By indirect observations, we mean that the kernels $(\theta^u)_{u\in {\cal U}}$ cannot be observed directly, but are revealed only through the trajectories $x^{\cal U}$. Most existing analysis  
\cite{ghosal1999posterior,petrone2012bayes,schwartz1965bayes} 
do not apply in our setting, as these papers aim at identifying conditions on the Bayesian prior $g$ and on the true distribution $\mu$ under which the Bayesian posterior $g|\theta^\set U$ will converge (either weakly or in $\text L_1$-norm) to $\mu$ in the limit of large population size. Hence, existing analysis are concerned with {\it direct} observations of the kernels $(\theta^u)_{u\in {\cal U}}$.

\section{Empirical Evaluation of CAMP}
\label{sec:empirical_evaluation}

\subsection{Mobility Traces}\label{subsec:sim}
We evaluate the performance of CAMP predictor using two sets of mobility traces collected on a Wi-Fi and cellular network, respectively.

{\noindent \bf Wi-Fi traces \cite{chon2013understanding}. } We use the dataset of \cite{chon2013understanding} where the mobility of 62 users are collected for three months in Wi-Fi networks mainly around a campus in South Korea. The smartphone of each users periodically scans its radio environment and gets a list of mac addresses of available access points (APs). To map these lists of APs collected over time to a set of locations, we compute the Jaccard index\footnote{Jaccard index between two lists $A$ and $B$ is defined as $\frac{|A \cap B|}{|A \cup B|}$.} between two lists of of APs scanned at different times. If two lists of APs have a Jaccard index higher than 0.5, these two lists are considered to correspond to a same geographical locations \cite{chon2013understanding}. From the constructed set of locations, we then construct the trajectories of the various users.

{\noindent \bf ISP traces \cite{de2014d4d}.} We also use the call detailed record (CDR) dataset provided by Orange where the mobility of 50000 subscribers in Senegal are measured over two weeks. We use the SET2 data~\cite{de2014d4d}, where the mobility of a given user is reported as a sequence of base station (BS) ids, and time stamps. Each record is obtained only when the user communicates with base stations (e.g., phone call, text message).

In each dataset, we first restrict our attention to a subset ${\cal L}$ of frequently visited locations. We select the 116 and 80 most visited locations in Wi-Fi traces and ISP traces datasets, respectively. We then re-construct users' trajectories by removing locations not in ${\cal L}$. For the ISP dataset, we extract 200 users (randomly chosen among users who visited at least 10 of the locations in ${\cal L}$). From the re-constructed trajectories, we observe a total number of transitions from one location to another equal to 8194 and 13453 for the Wi-Fi and ISP dataset.


\begin{figure}[]
  \centering
  \subfigure[Wi-Fi traces]{
  \includegraphics [width=0.88\columnwidth ]{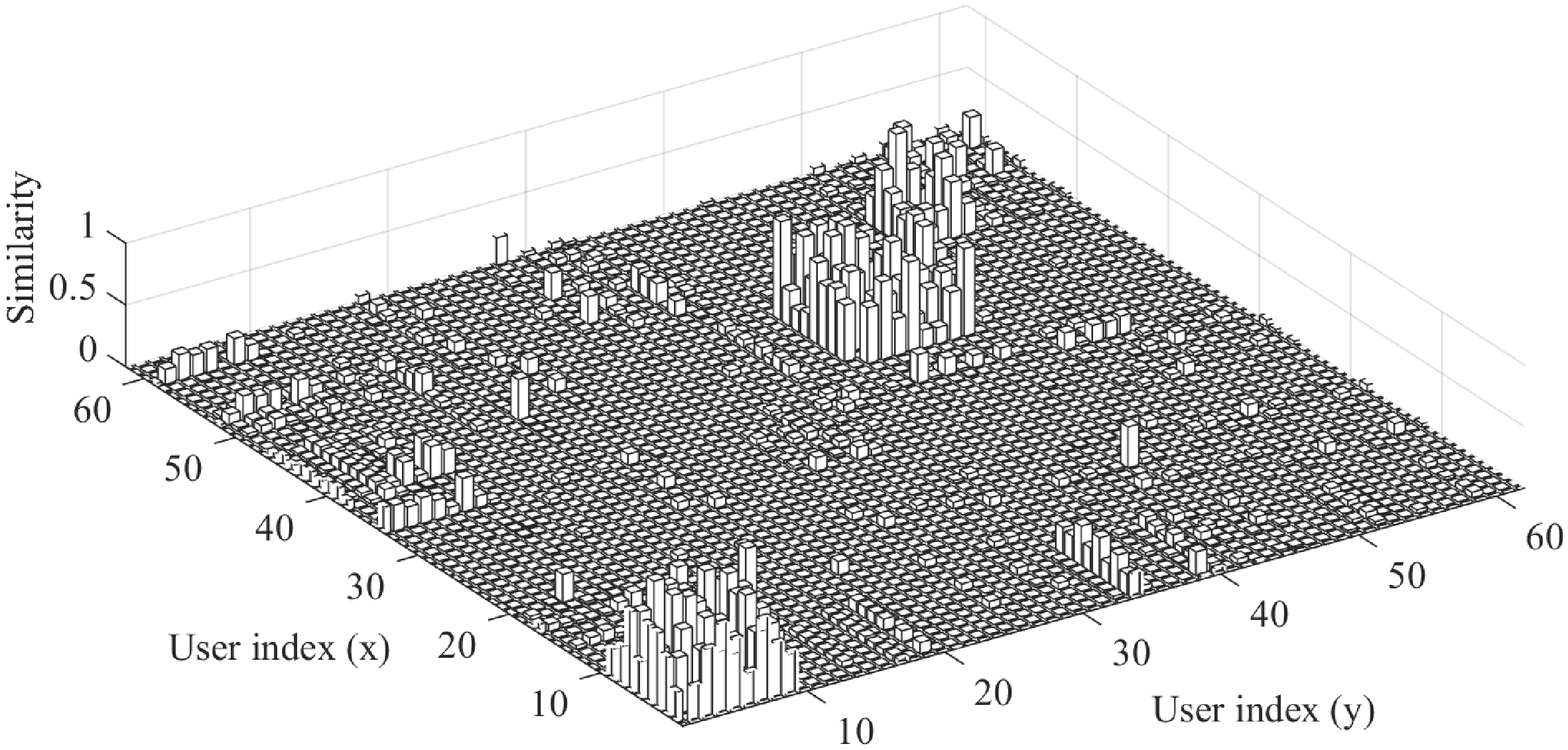}}
  \subfigure[ISP traces]{
  \includegraphics [width=0.88\columnwidth ]{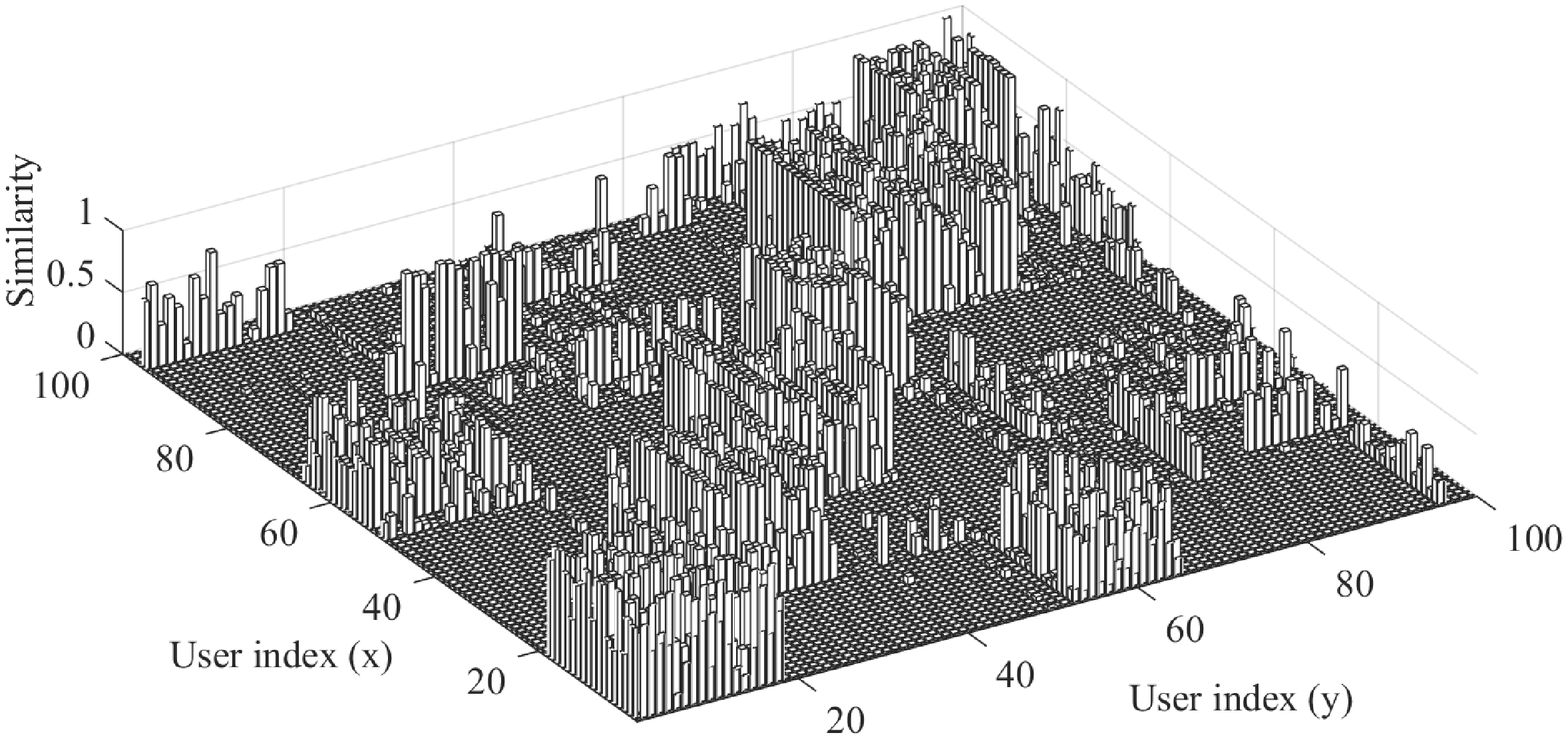}}
  \caption{Similarities between pairs of users. For the ISP traces we restrict the plot to 100 randomly selected users.}
\label{fig:similarity_map}
\end{figure}

{\noindent \bf Users' similarity.} Before actually evaluating the performance of various prediction algorithms, we wished to assess whether users exhibit similar mobility patterns, that could in turn be exploited in our predictions. Here, we test the similarity of pairs of users only. More precisely, we wish to know whether the observed trajectory of user $v$ could be aggregated to that of user $u$ to improve the prediction of user-$u$'s mobility. To this aim, we use the concept of mutual prediction~\cite{schiff1996detecting} as follows. 

We first define the empirical accuracy of an estimator $\hat\theta$ of user-$u$'s transition kernel:  
\begin{eqnarray}
\overline{AC}^u(\hat \theta) = \frac{1}{n^u-1}\sum_{t = 2}^{n^u} \ind(x^u_t = \arg \max_j \hat \theta_{x^u_{t-1},j}) 
\label{eq:emp_pr}
\end{eqnarray}
Let $\hat\theta^{u*}$ be the maximum likelihood estimator of $\theta^u$ given $x^u$ (i.e., $\hat\theta^{u*}_{i,j} = \frac{n^u_{i,j}}{n^u_i}, \forall i,j \in \set L$ ). Intuitively, user-$v$'s trajectory is useful to predict the mobility of user $u$ if $\hat\theta^{v*}$ has a high empirical accuracy for user $u$, i.e., if $\overline{AC}^u(\hat\theta^{v*})$ is high. We hence define the {\it similarity} $sim(u,v)$ of users $u$ and $v$ as $sim(u,v)= \overline{AC}^u(\hat\theta^{v*}) / \overline{AC}^u(\hat\theta^{u*})$.
Note that the notion of similarity is not symmetric (in general $sim(u,v)\neq sim(v,u)$), and it always takes its value between 0 and 1. 

Fig.~\ref{fig:similarity_map} (a) and (b) present the similarity between 62 users in Wi-Fi trace and 100 users in the ISP subscriber dataset. To provide meaningful plots, we have ordered users so that pairs of users with high similarity are neighbours (to this aim, we have run the spectral clustering algorithm~\cite{jebara2007spectral} and re-grouped users in the identified clusters). From these plots, the similarity of users is apparent, however we also clearly observe that perfect clusters (in which users' patterns are exactly same) do not really exist. 
From the dataset, we observe that 1.65\% and 5\% of user pairs out of all possible pairs have similarity higher than 0.5 for the Wi-Fi and ISP traces. We also computed the number of users having at least one user with whom the similarity is higher than 0.5. In the Wi-Fi traces, we found 19 (out of 62) such users, whereas in the ISP traces there are 173 (out of 200) such users. These numbers are high, and justify the design of cluster-aided predictors.

\subsection{Prediction Accuracy}
\subsubsection{Tested Predictors} 
We assess the performance of six types of predictors: the order-1 Markov predictor (Markov~\cite{song2006evaluating}), the order-2 Markov predictor with fallback (Markov-O(2)~\cite{song2006evaluating}), AGG, CAMP and CAMP$^C$, AGG$^C$.
Before describing each predictor, we briefly introduce some notations regarding the training data available at a given time. The time stamp of the arrival at $t$-th location on user-$u$'s trajectory is denoted by $d^u_t \in \R$, and $n^u(d)$ is the length of user-$u$'s trajectory collected before time $d$ (i.e., $n^u(d) = \max \{ s |d^u_s < d\}$). The collection of users' trajectories available for a prediction at time $d$ is denoted by $x^{\set U,d}$  (i.e., $x^{\set U,d} = ( x^{v,d} )_{v \in \set U},$ where $x^{v,d} = (x^v_1,..,x^v_{n^v(d)})$). The prediction for $x^u_t$ is denoted by $\hat x^u_t.$

In order to derive an estimate of the $t$-th location $\hat x^u_t$ of user $u$, the Markov predictors first estimate $\theta^u$ based on user-$u$ trajectory only, i.e., based on $x^{u,d^u_t}$. In contrast, AGG and CAMP algorithms exploit the data available on all users $x^{\set U,d^u_t}$ to estimate $\theta^u$. The AGG algorithm tries in a very naive way to exploit users' similarities. It considers that all users have the same transition kernel (as if there were a single cluster only), and thus uses all trajectories (in the same way) to estimate $\theta_u$. CAMP$^C$ (resp. AGG$^C$) differs from CAMP (resp. AGG) in that its prediction at time $d$ under for user $u$ uses other users' complete trajectories (i.e., $x^{\set U \setminus u}$). This corresponds to a case where user $u$ starts moving along her trajectory after other users have gathered sufficiently long trajectories. Under all algorithms, the estimated $\theta^u$ is denoted by $\hat\theta^{u,d^u_t}$). Finally, Markov-O(2) assumes that users' trajectories are order-2 Markov chains, and for the locations where the corresponding order-2 transitions are not observed, Markov-O(2) falls back to the Markov predictor. The description of the various predictors is summarized in Table 1.

\begin{table}[]
\tabcolsep = 0.07in
{
\caption{Order-1 predictors.} 
\begin{tabular}{|c||c|c|c|} \hline 
    & $\hat\theta^{u,d^u_t}$ & $\hat x^u_t$   \\ \hline \hline
  \sl Markov~\cite{song2006evaluating} & $\arg\max_{\theta} P_\theta(x^{u,d^u_t})$ &\multirow{5}{*}{$\arg \max_j \hat\theta^{u,d^u_t}_{x^u_{t-1},j}$}  \\ \cline{1-2}
  \sl AGG & $\arg\max_{\theta} P_\theta(x^{\set U,d^u_t})$ &    \\ \cline{1-2}
  \sl CAMP & $\simeq E_g[\theta^u | x^{\set U,d^u_t}]$ &    \\ \cline{1-2}
  \sl AGG$^C$ & $\arg\max_{\theta} P_\theta(x^{\set U \setminus u},x^{u,d^u_t})$ &    \\ \cline{1-2}
  \sl CAMP$^C$ & $\simeq E_g[\theta^u | x^{\set U \setminus u},x^{u,d^u_t}]$ &    \\  \hline
\end{tabular}
}
\end{table}

The parameters $B$, $K$ and $M$ for CAMP and CAMP$^C$ are set to 8, 3 and 30.

\begin{figure*}[]
  \centering
  \subfigure[CAPR$_{time}$, Wi-Fi traces]{
  \includegraphics [width=0.242\textwidth]{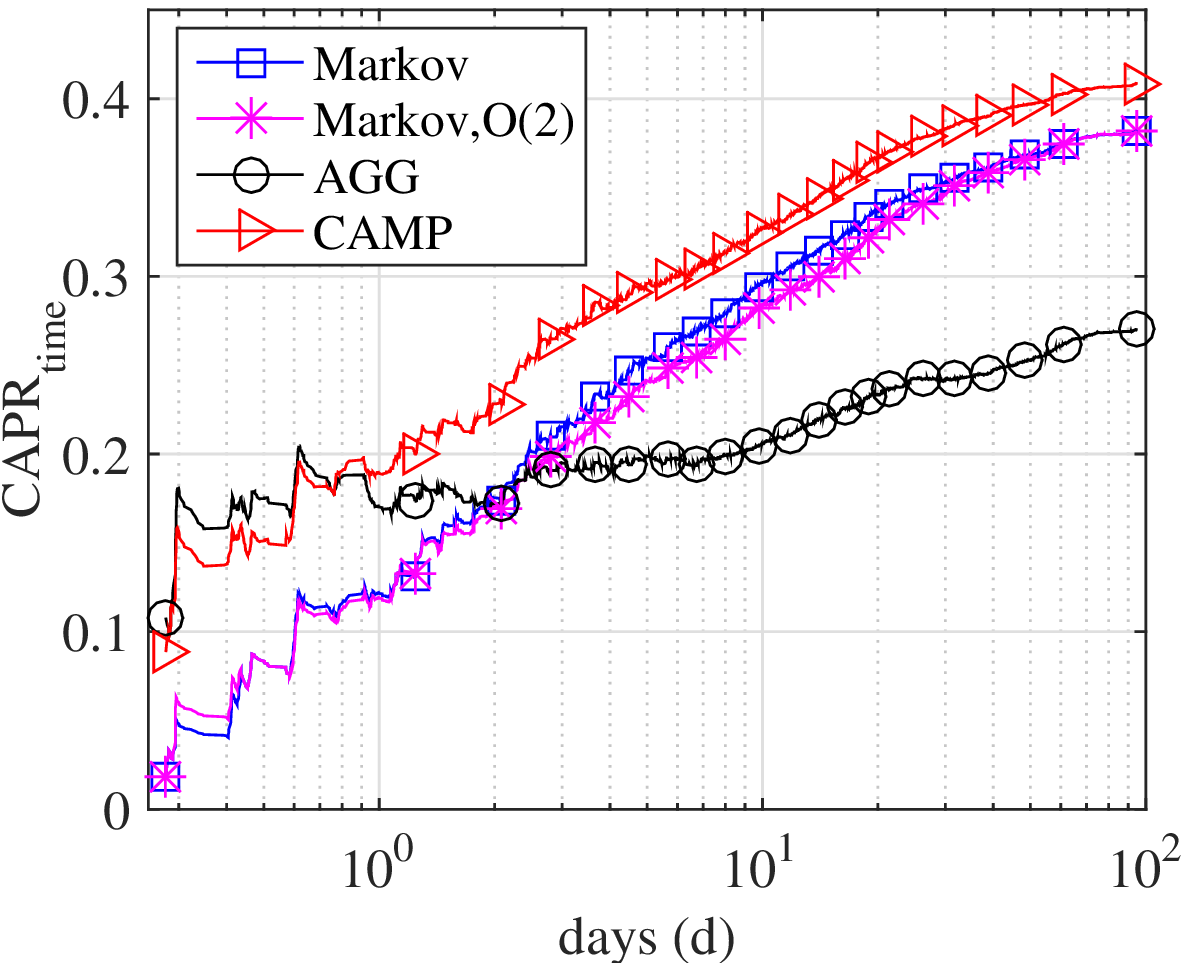}}
  \subfigure[CAPR$_{time}$, ISP traces]{
  \includegraphics [width=0.242\textwidth]{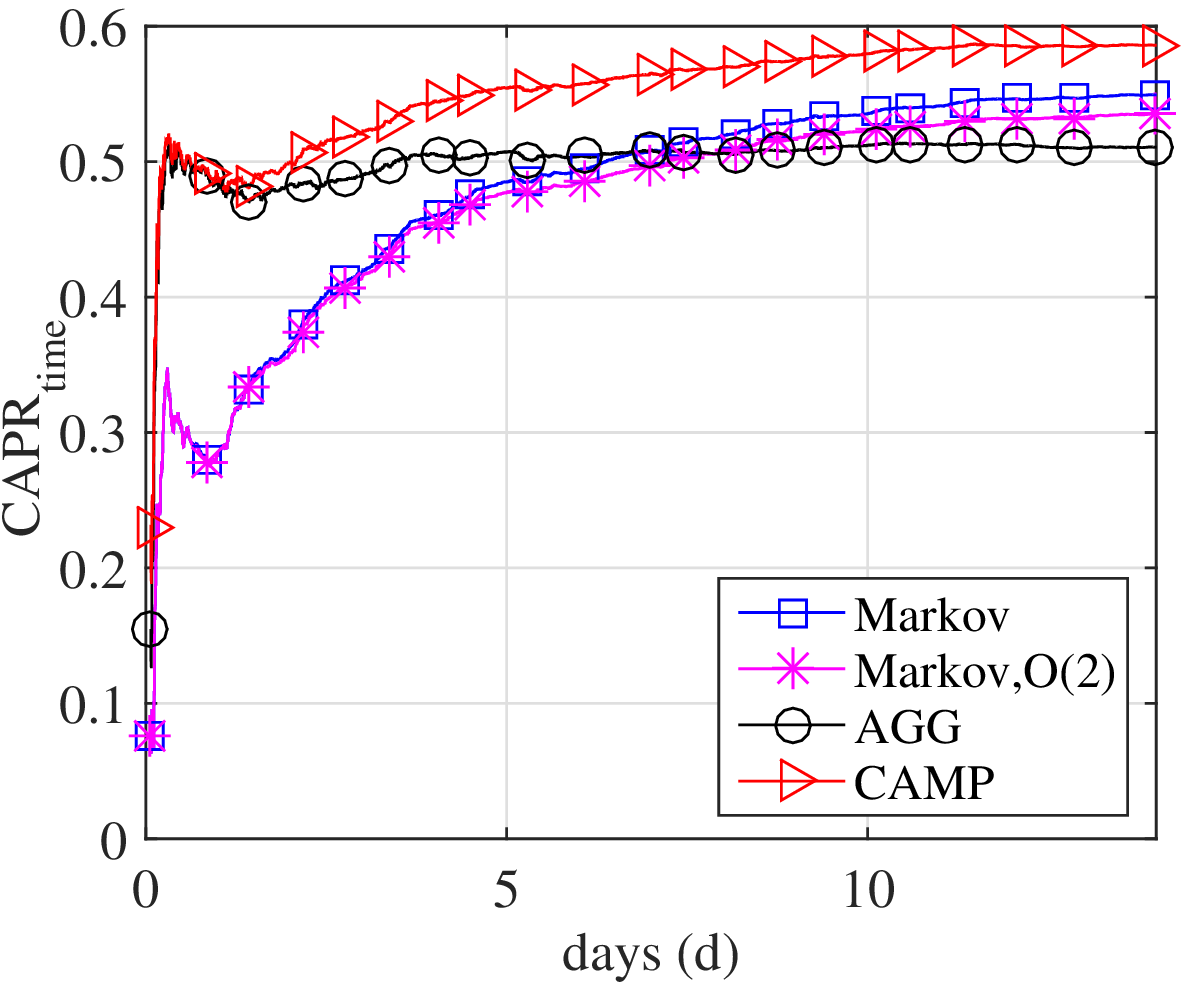}}
  \subfigure[CAPR, Wi-Fi traces]
  {\includegraphics [width=0.242\textwidth ]{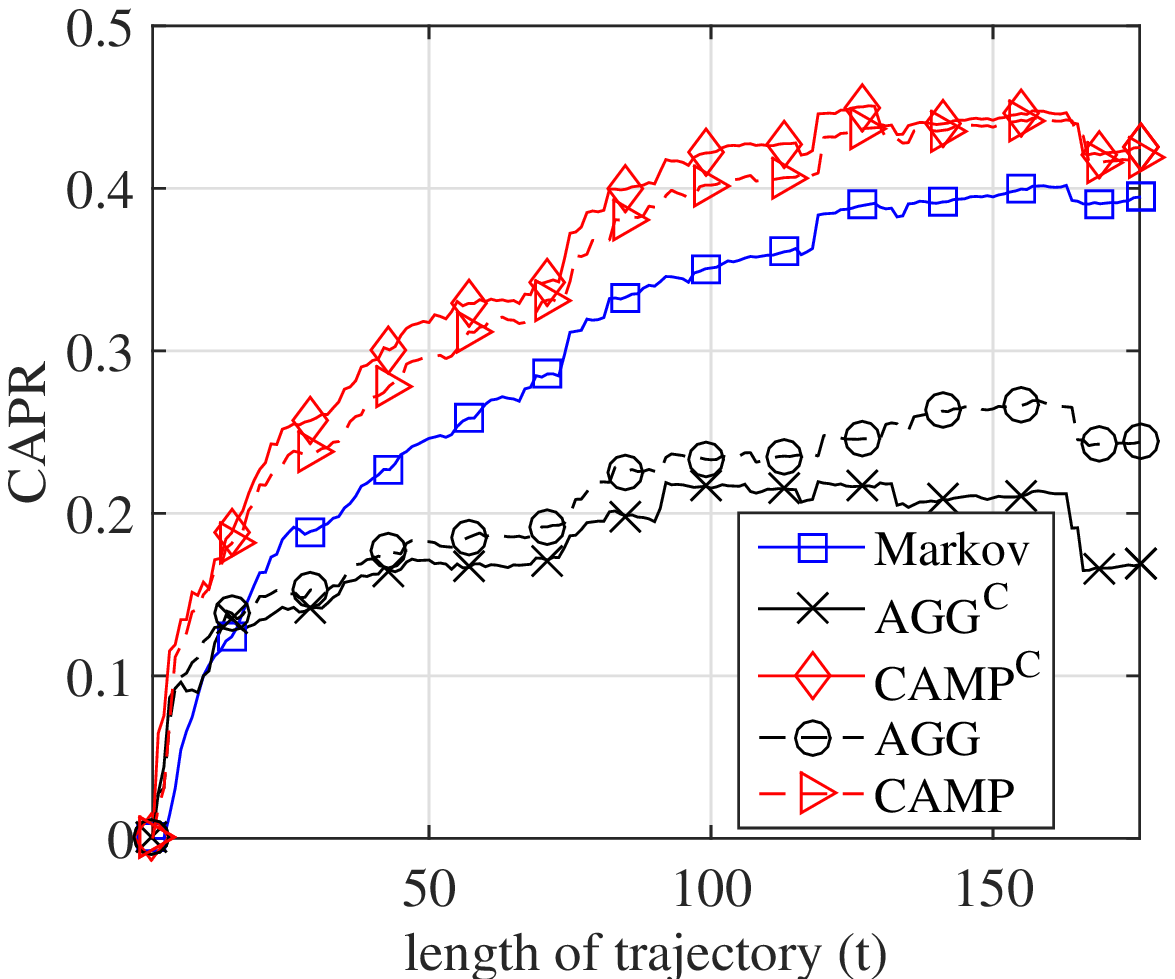}} 
  \subfigure[CAPR, ISP traces]
  {\includegraphics [width=0.242\textwidth ]{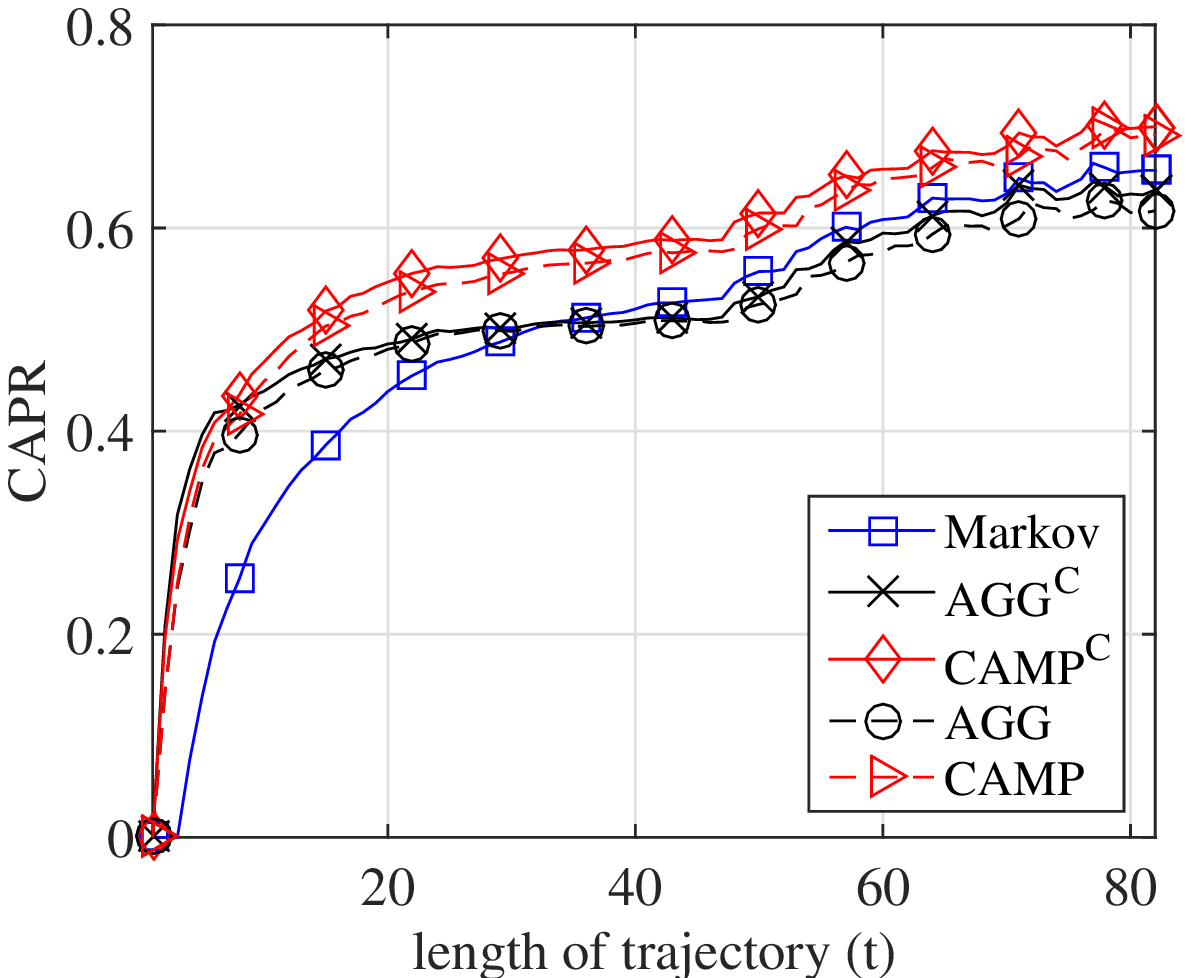}}  \\
  \subfigure[IAPR, Wi-Fi traces]
  {\includegraphics [width=0.244\textwidth ]{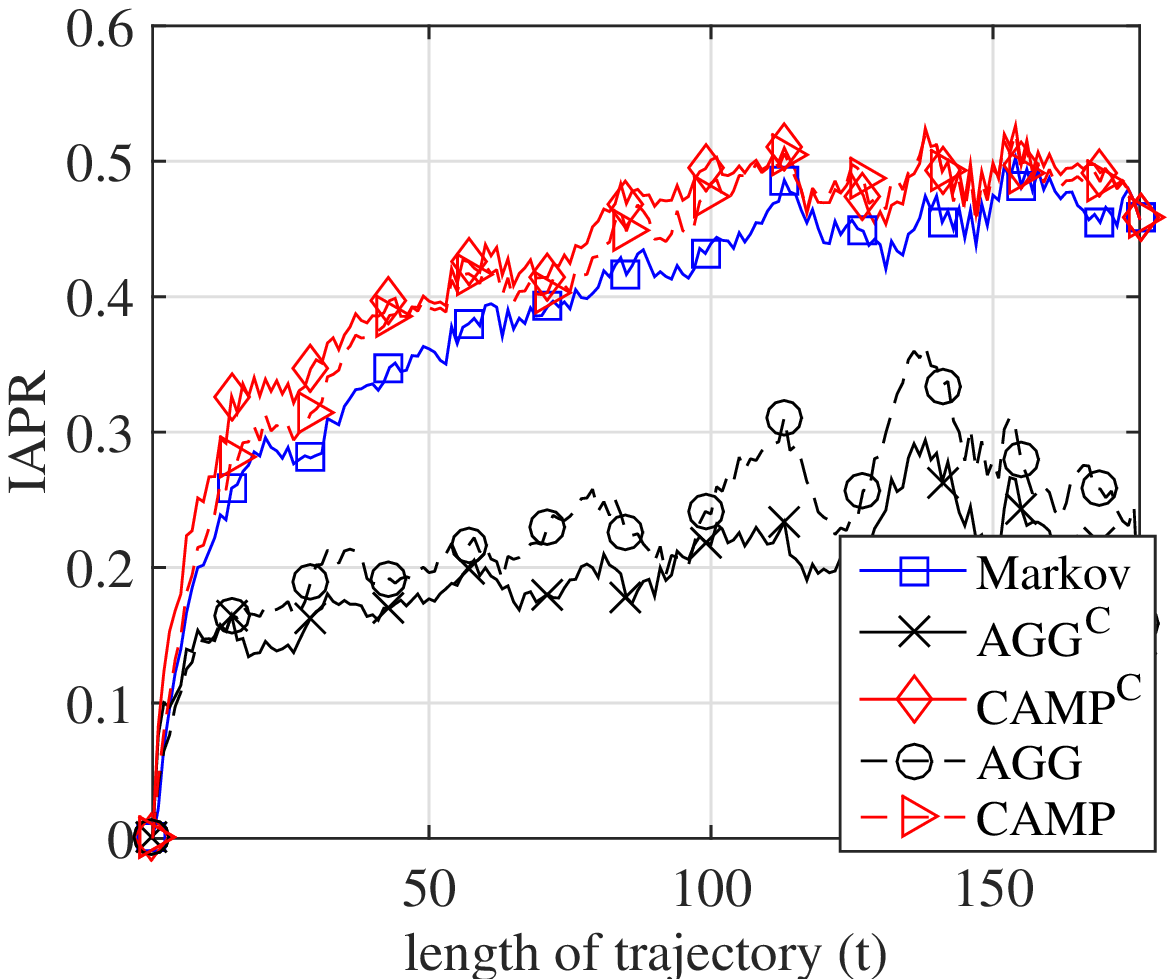}} 
  \subfigure[IAPR, ISP traces]
  {\includegraphics [width=0.244\textwidth ]{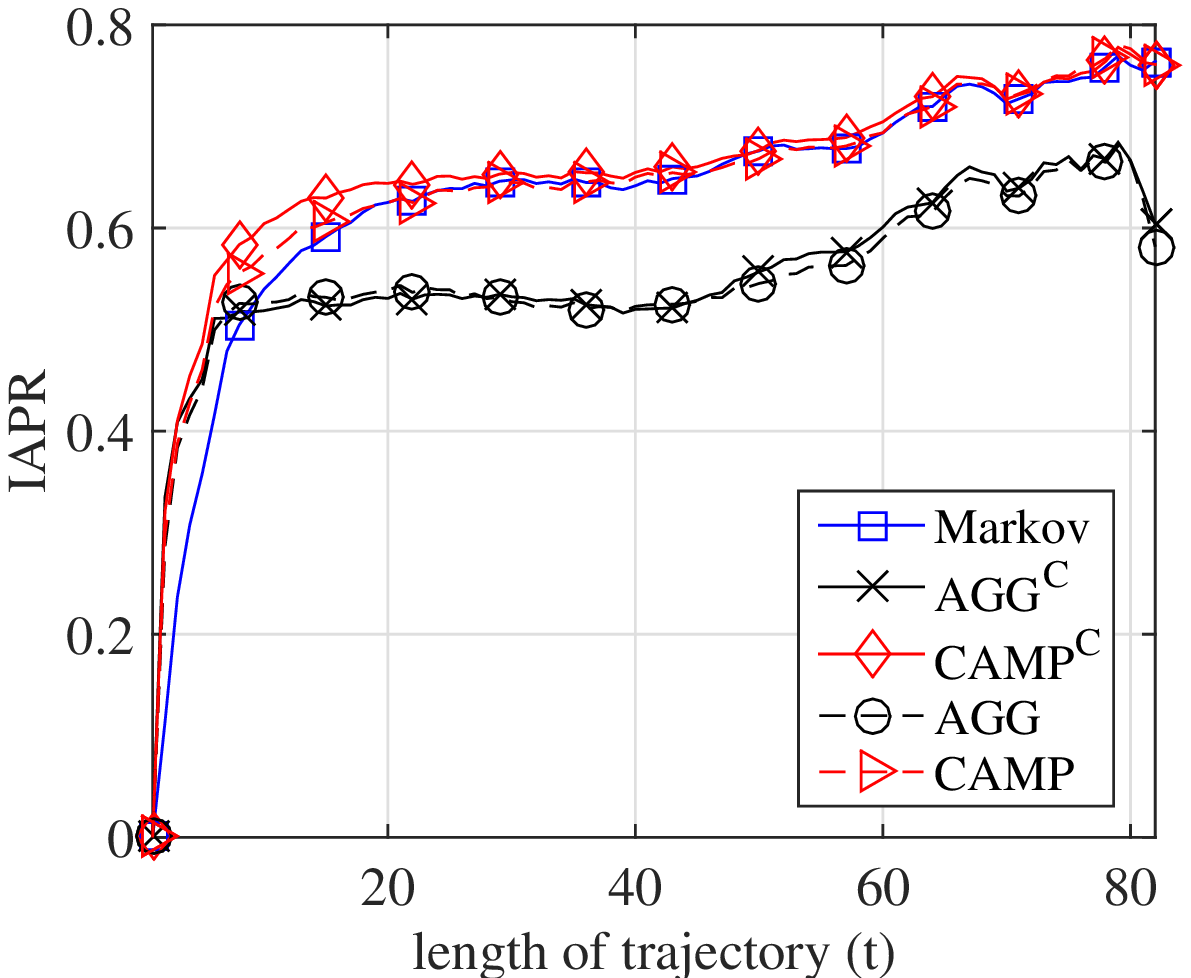}}
  \subfigure[CAPR, Wi-Fi traces (MF)]
  {\includegraphics [width=0.244\textwidth ]{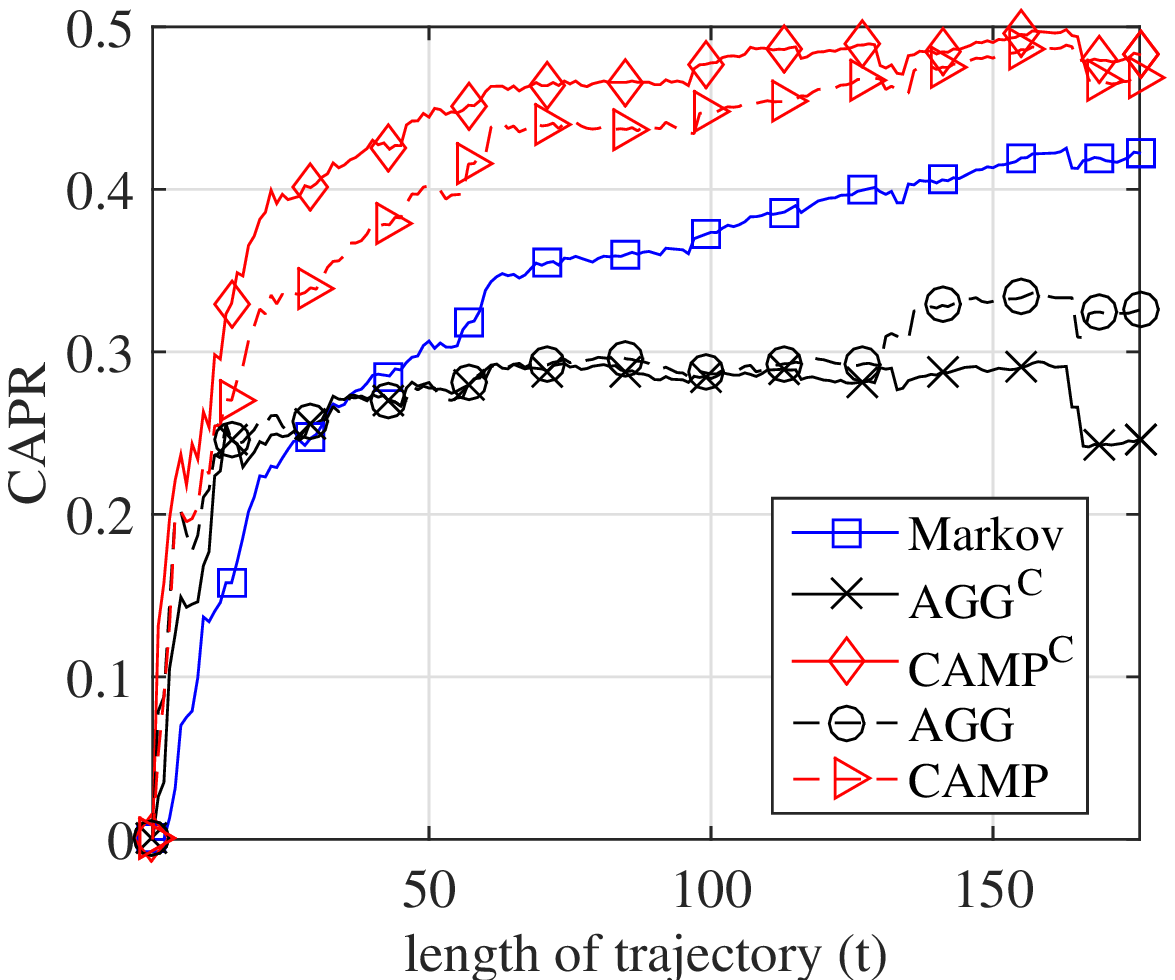}}
  \subfigure[IAPR, Wi-Fi traces (MF)]
  {\includegraphics [width=0.244\textwidth ]{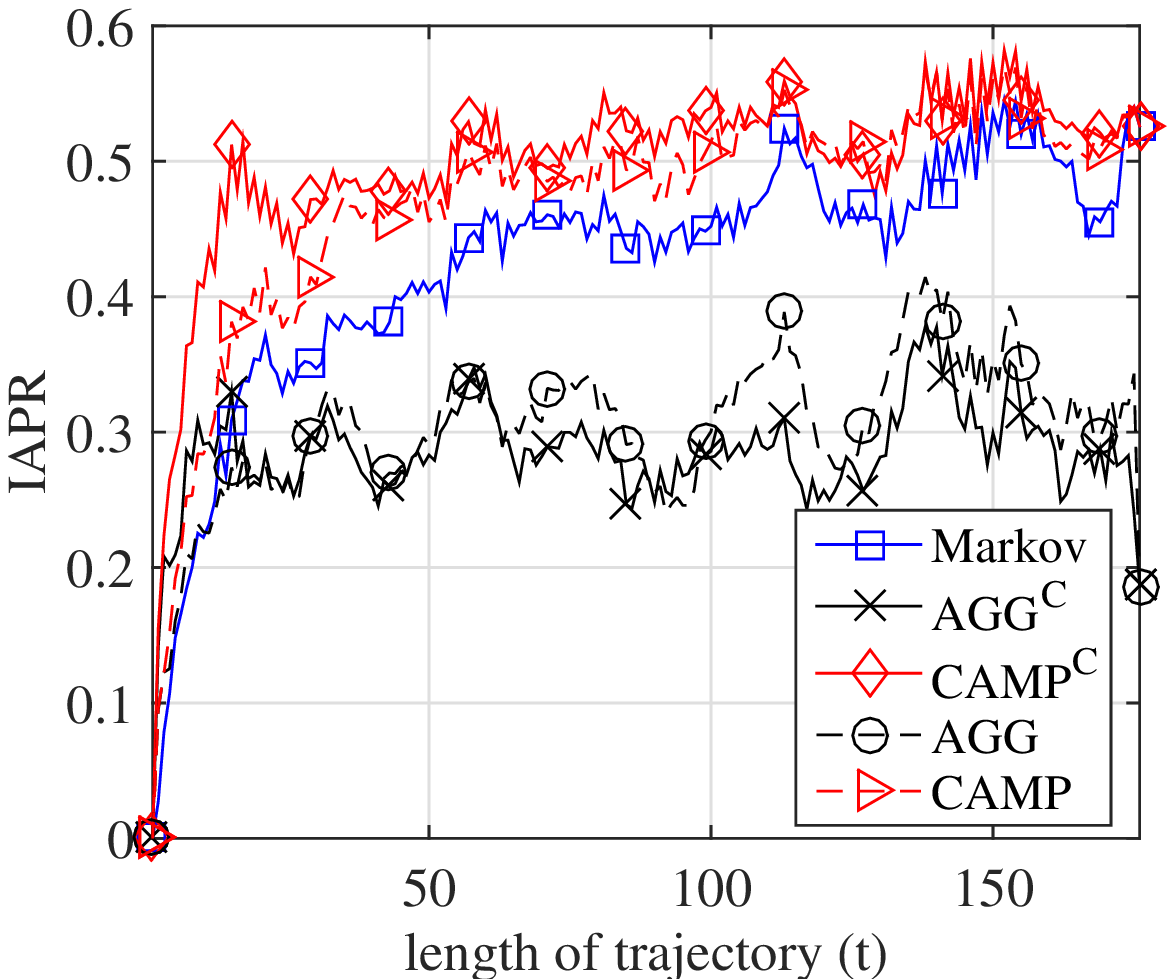}} 
\caption{Performance of various predictors. 
\label{fig:carar}}
\vspace{-0.3cm}
\end{figure*}

%

\subsubsection{Results}  

We assess the performance of the various algorithms using two main types of metrics. The first metric, referred to as the Cumulative Accurate Prediction Ratio (CAPR), is defined as the fraction of accurate predictions for all users up to time $d$:
$$
CAPR_{time} = \frac{1}{\sum_{u\in \set U} (n^u(d)-1) }\sum_{u \in \set U}\sum_{s=2}^{n^u(d)} \ind(\hat x^u_s =  x^u_{s}).
$$
We also introduce a similar metric that captures the cumulative accuracy of predictions after observing $t$ different locations on users' trajectories: 
$$
CAPR = \frac{1}{(t-1)\sum_{u \in \set U} \ind(n^u \geq t)} \sum_{\substack{u \in \set U \\ n^u \geq t}} \sum_{s=2}^t \ind(\hat x^u_s =  x^u_{s}).
$$
The second type of metrics concerns the instantaneous accuracy of the predictions. The Instantaneous Accurate Prediction Ratio (IAPR) after observing $t$ different locations on users' trajectories is defined as follows. 
$$
IAPR = \frac{1}{\sum_{u \in \set U} \ind(n^u \geq t)} \sum_{u \in \set U, n^u \geq t} {n^u_{x^u_{t-1},\hat x^u_t}\over n^u_{x^u_{t-1}}}.
$$

Fig.\ref{fig:carar}(a)-(b) present $CAPR_{time}$ as a function of time $d$ for various algorithms and for the two mobility traces. CAMP outperforms all other algorithms at any time. The improvement over Markov and Markov-O(2) can be as high as 65\%. This illustrates the performance gain that can be achieved when exploiting users' similarities. Note Markov-O(2) does not outperform Markov, which was also observed in \cite{chon2012evaluating}. In the following, we only evaluate the performance of the Markov predictor, and do not report that of its order-2 equivalent.

In Fig.\ref{fig:carar} (c)-(f), we plot the CAPR and IAPR as a function of the length $t$ of the observed trajectory. In Fig.\ref{fig:carar}(c) and (d), when the collected trajectory is not sufficient (i.e., $t=10$), CAMP$^C$ and CAMP outperforms Markov by 64\% and 40\%, respectively. Regarding the IAPR in Wi-Fi traces, Fig~\ref{fig:carar}(e) shows that CAMP and CAMP$^C$ provide much better predictions than Markov, when the length of trajectory is less than 140. After a sufficient training data is collected, they yield comparable IAPR. In Fig~\ref{fig:carar} (f), for the ISP traces, the IAPR under CAMP and Markov are similar sooner, for trajectories of length greater than 20 only. 

In Fig.\ref{fig:carar} (g) and (h), we evaluate the CAPR and IAPR averaged only over users having at least one user with whom the similarity is higher than 0.5 (see \S\ref{subsec:sim}). These users are referred to as Mobility Friendly (MF) users. In Fig.\ref{fig:carar}(g), we observe that for MF users, the gain of CAMP$^C$ and CAMP becomes really significant, i.e., when $t$=10, the CAPR of CAMP$^C$ and CAMP outperform that of Markov by 102\% and 65\%, respectively. Also note that CAMP$^C$ becomes significantly better than CAMP for MF users. This is explained by the fact that we can predict the mobility of MF users much more accurately if we have a long history of the mobility of users they are similar to. The performance for MF users in the ISP traces is not presented, because there, most of users (i.e., 86\%) are already MF users.


\subsubsection{Exploiting Similarities in CAMP} \label{sec: resemble}


Recall that, by the weight of the empirical transition kernel of user $v$ (i.e., $\gamma^v_i$) in computing $\hat \theta^u$ in \eqref{eq: hat theta resem}, we can quantify to what extent the observed trajectory of user $v$ is taken into account in the estimate $\hat\theta^u$. When summing $\gamma^v_i$ over all locations $i$, we get an aggregate indicator capturing how $v$ impacts the prediction for user-$u$'s mobility. To understand how many users actually impact the prediction for user $u$ in the CAMP, we may look at the cardinality of the set of users whose aggregate indicator exceeds a given threshold:   
$ \{v |z \sum_{i \in \set L} \gamma^v_i > \frac{1}{|\set U|}\},$
where $z$ is a normalization constant to make the sum of aggregate indicators over all users equal to 1. The above set is called the set of $u$-similar users.

\begin{figure}[]
  \centering
  \subfigure[Wi-Fi traces]
  {\includegraphics [width=0.49\columnwidth ]{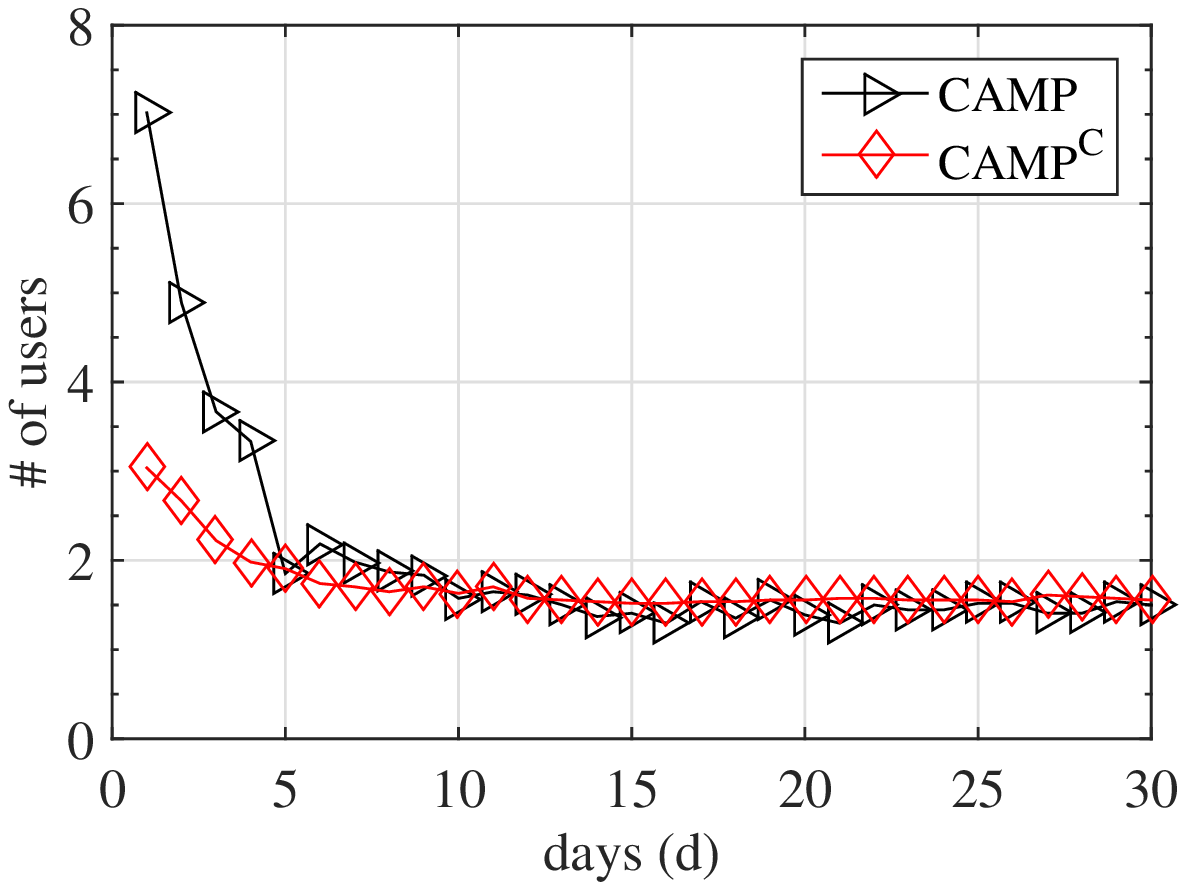}} 
    \subfigure[ISP traces]
  {\includegraphics [width=0.49\columnwidth ]{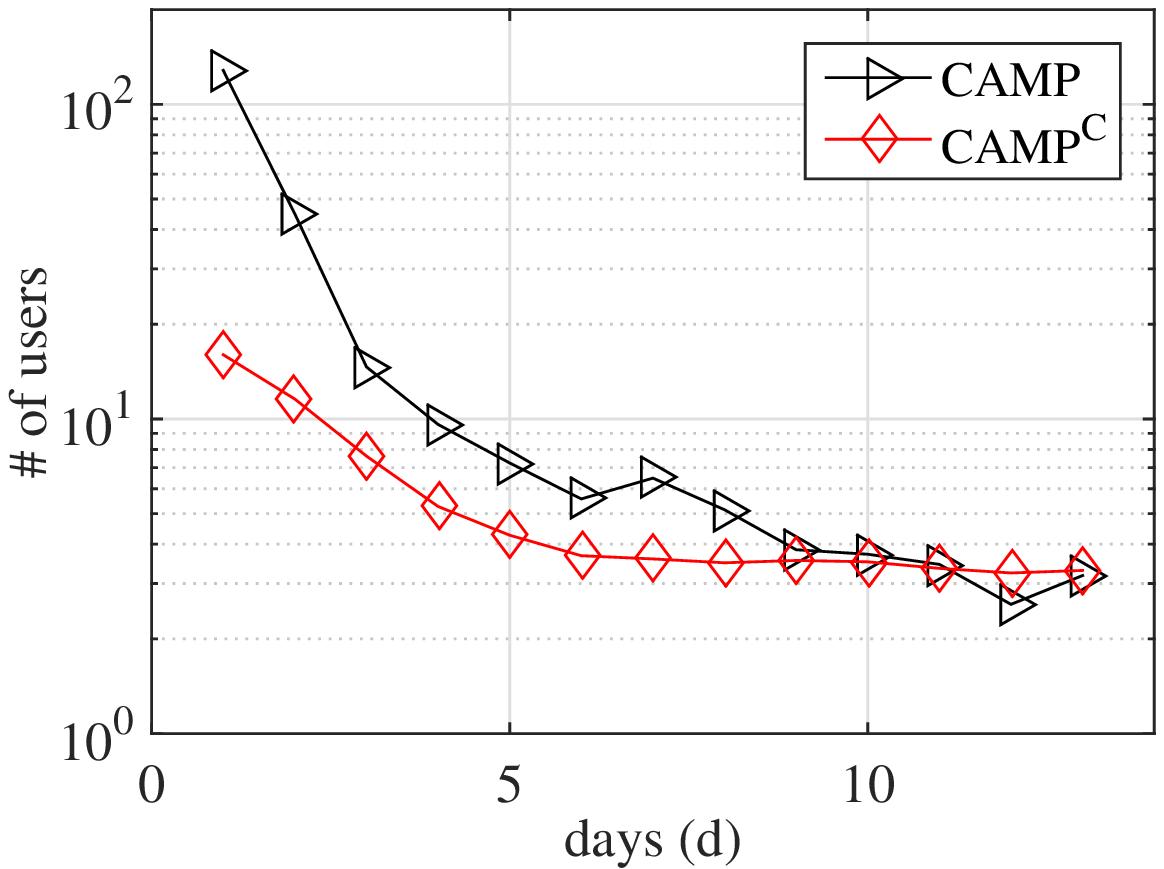}} 
  \caption{Number of $u$-similar users, averaged over all users $u$, vs. time.}
  \label{fig:count_resem}
\end{figure}

In Fig.\ref{fig:count_resem}, we plot the number of $u$-similar users, averaged over all users $u$, and as a function of the length of trajectories (in days $d$). In case of CAMP, the first day, the average numbers are 7 and 110 in Wi-Fi traces and ISP traces, which means that CAMP aggressively uses the trajectories of all users for its prediction. 
When the length of the trajectories increase, the average size decreases to 1.5 after one month in Wi-Fi traces and 2.2 after two weeks in ISP traces. In other words, as data is accumulated, CAMP does not use the trajectories of a lot of users for its prediction. This illustrates the adaptive nature of CAMP, which only exploits similarities among users if this is needed. In the case of CAMP$^C$, we observe a faster decrease with time of the average number of $u$-similar users, which means that CAMP$^C$ tends to utilize other users' data more selectively, even at the beginning. This explains why CAMP$^C$ performs better than CAMP in Fig.\ref{fig:carar}.

\begin{figure}[]
  \subfigure[Distribution of estimation errors]
  {\includegraphics [width=0.238\textwidth ]{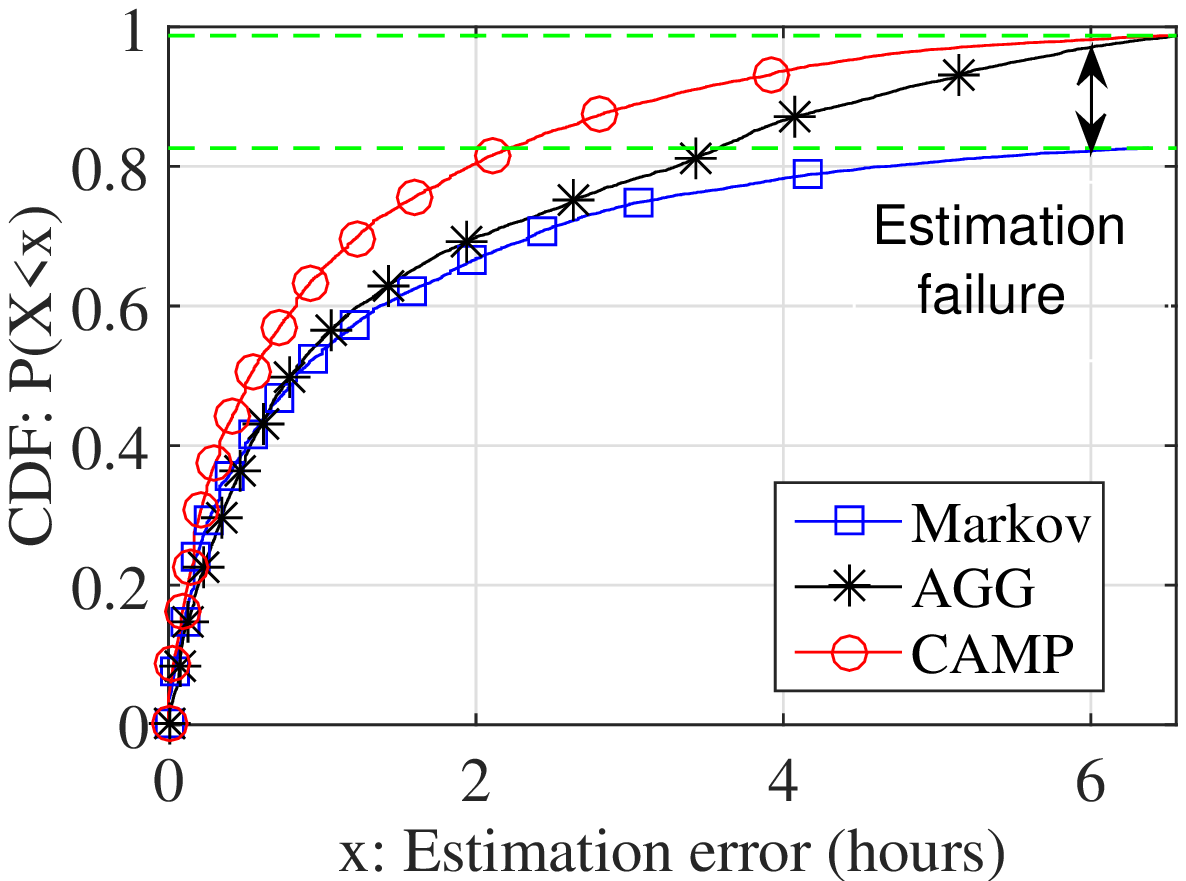}}  
  \subfigure[When Markov is unavailable]
  {\includegraphics [width=0.238\textwidth ]{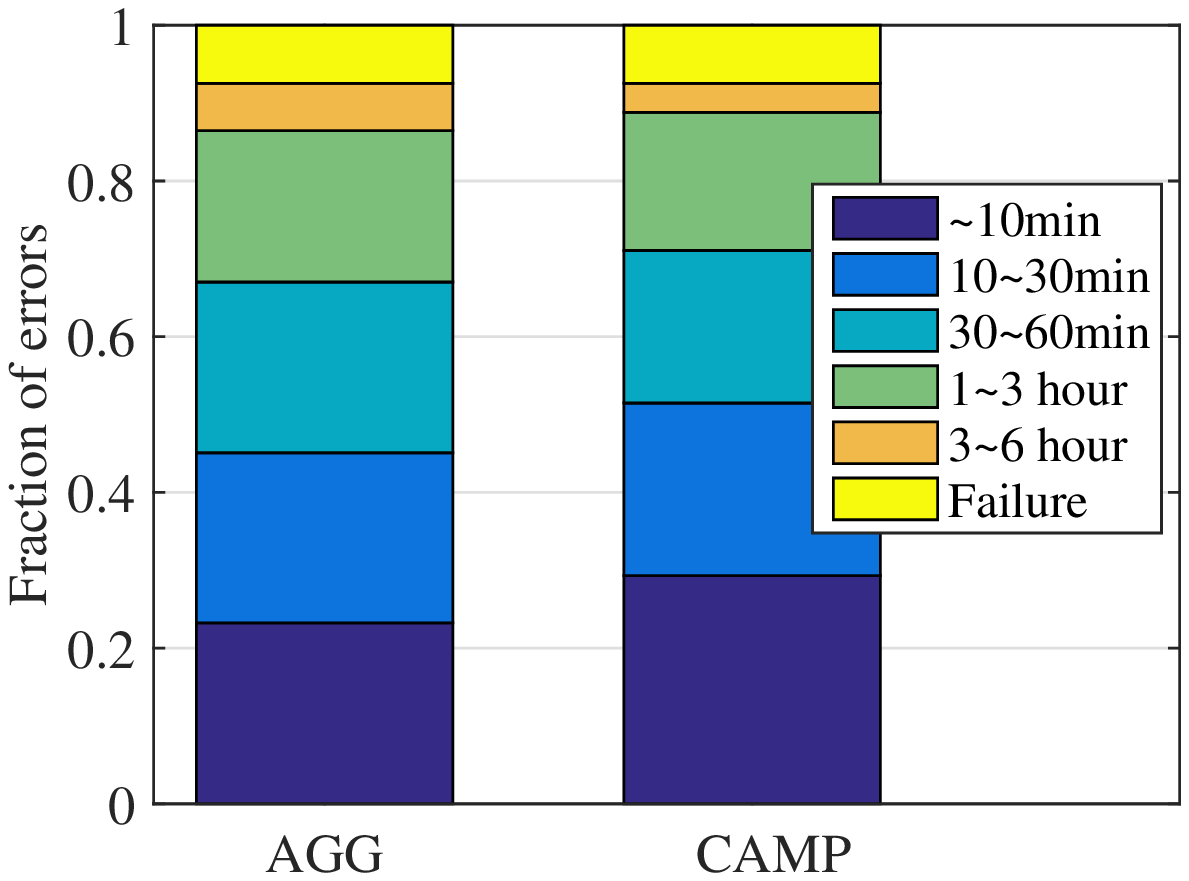}}  
  \caption{Estimation error of staying time in Wi-Fi trace. The dashed lines in (a) indicate the fraction of cases where the related training data was collected. \label{fig:stay_time}}
\end{figure}

\subsubsection{Error of Staying Time Estimation}
\label{sec: error_staying_time}

In our scenario, where each user $u$ arrives at $t$-th location $x^u_t$, a predictor estimates the staying time $\hat s^u_t$ with the available data.
Markov predictor~\cite{chon2012evaluating,scellato2011nextplace}(resp. AGG) computes the average of staying times of user $u$ (resp. all users) which have been measured at the location $x^u_t$ until $d^u_t$. CAMP predicts $\hat s^u_t$ by computing the equation~\eqref{eq: hat_stay} with the observed data of all users. 
The performance metric for each user $u$ measured at $t$-th location is the difference between the estimated and acutual staying time ( i.e., $|\hat s^u_t - s^u_t|$). We call it as {\em estimation error}. We test the estimation error only with Wi-Fi trace, because we cannot precisely observe staying time in ISP trace in which a location is recorded not periodically, but only when users randomly communicate with base stations.

Fig.\ref{fig:stay_time} (a) plots CDFs of estimation errors of every user $u$ and $t$ obtained by tested predictors. 
CAMP provides lower estimation error than that of Markov and AGG. The median of CAMP is less than those of Markov and AGG by 35\% and 28\%, respectively. For 18\% of all instances (marked as ``Estimation failure"), Markov couldn't provide estimations, because the individual users haven't collected their staying times at the current location before. However in those cases AGG and CAMP are still able to estimate the staying time by using other users' observations. In Fig.\ref{fig:stay_time} (b), we further test the estimation quality of AGG and CAMP, when Markov is unavailable due to lack of the individual training data. In that case, 43\% of estimations provided by CAMP give less than 30 minutes errors. Median of estimation errors of CAMP is 13.4\% less than that of AGG, because CAMP selectively utilizes other users' data.

\section{Concluding Remarks}
%
%
In this paper, we have presented a cluster-aided inference method to predict the mobility of users in wireless networks. This method significantly departs from existing prediction techniques, as it aims at exploiting similarities in the mobility patterns of the various users to improve the prediction accuracy. The proposed algorithm, CAMP, relies on Bayesian non-parametric estimation tools, and is robust and adaptive in the sense that it exploits users' mobility similarities only if the latter really exist. We have shown that our Bayesian prediction framework can asymptotically achieve the performance of an optimal predictor when the user population grows large, and have presented extensive experiments indicating that CAMP outperforms any other existing prediction algorithms. Note also that CAMP can be implemented without damaging users' privacy (the data can be anonymized).
 
Many interesting questions remain about the design of CAMP. In particular, we plan to investigate how to set its parameters ($B$, $K$, and $M$) to achieve an appropriate trade-off between accuracy and complexity. These parameters could also be modified in an online manner while the algorithm is running to adapt to the nature of the data.  We further plan to apply the techniques developed in this paper to various kind of {\it mobility}, e.g., we could investigate how users dynamically browse the web, and use our framework to predict the next visited webpage.  

\section*{Appendix}

\subsection{Proof of Lemma~\ref{lem: computation}}

Observe that in view of (\ref{eq:update G_0}), we have:
\begin{equation}
G_0^{k+1}(\ud\theta)=\sum_c\omega_c^k P_\theta(x^c)G_0^k(\ud\theta),
\label{eq: recur_updateG0}
\end{equation}
where the sum is over all possible partitions of the set of users $\set U$ in clusters and the weight $\omega_c^k$ is
\begin{equation}
\omega_c^k=\frac{n_c^k}{B|\set U|\int_\theta P_\theta(x^c)G_0^k(\ud\theta)}, \label{eq: omega}
\end{equation}
with $n_c^k=\sum_{b=1}^B\sum_{u\in\set U}\ind(c^{u,b,k}=c)$. 
Recursively replacing $G_0^k$ in \eqref{eq: recur_updateG0} with $G_0^{k-1}$ and putting $G^1_0$ = Uniform($\Theta$),  
we obtain another expression of $G_0^K$ as 
\begin{align}
G_0^{K}(\ud\theta)=\sum\limits_{c_1,\ldots,c_{K-1}} \prod\limits_{k=1}^{K-1} \omega_{c_k}^k P_\theta(x^{c_k})\ud\theta,\label{eq: G_0 K+1}
\end{align}
where the sum $\sum\limits_{c_1,\ldots,c_{K-1}}$ is $\sum_{c_1 \in \set C_1}\cdots \sum_{c_{K-1} \in \set C_{K-1}}$ where 
$\set C_k$ is a set of every cluster sampled at $k$-th iterations, i.e., $\set C_k = \{ c | \sum_{b=1}^B \sum_{u \in \set U} \ind(c^{u,b,k}=c) > 0 \}$.
We can further obtain the recursive expression of the weights $\omega_c^K$ by plugging \eqref{eq: G_0 K+1} in \eqref{eq: omega}:
\begin{eqnarray}\label{eq:final}
\omega_c^{K}=\frac{n_c^{K}}{B|\set U|  \sum\limits_{c_1,\ldots,c_{K-1}}\xi_{c_1,\ldots,c_{K-1},c} \prod\limits_{k=1}^{K-1}\omega_{c_k}^k } ,
\end{eqnarray} 
\begin{eqnarray}
\xi_{c_1,\ldots,c_K}&=&\int_\theta\prod_{k=1..K}P_\theta(x^{c_k})\ud\theta \label{eq:xi}\\
&=&\prod_{i\in\set L}\frac{\prod_{j\in\set L}\Gamma(1+\sum_{k=1..K}n_{i,j}^{c_k})}{\Gamma(|\set L|+\sum_{k=1..K}n_i^{c_k})}, 
\end{eqnarray}
where $n^c_{i,j}=\sum_{u\in c}n^u_{i,j}$, $n^c_i=\sum_{j\in\set L}n^c_{i,j}$. 

Then, using equations~\eqref{eq:theta_com}, \eqref{eq: G_0 K+1} and \eqref{eq:xi}, we get an expression for $\hat\theta^{u}_{i,j}$:
In \eqref{eq:theta_com}, replacing the denominator of each sample $b$ with  $\omega^K_{c^{u,b,K}},$ and plugging \eqref{eq: G_0 K+1} into numerator,
we arrive at
\begin{eqnarray}
\hat\theta^{u}_{i,j}= \frac{1}{B}\sum_{b=1}^B \frac{\omega_{c^{u,b,K}}^KB|\set U|}{n^K_{c^{u,b,K}}} \sum\limits_{c_1,\ldots,c_{K-1}} \int_\theta \theta_{i,j}P_\theta(x^{c^{u,b,K}})  \cr
 \prod\limits_{k=1}^{K-1} P_\theta(x^{c_k}) \omega_{c_k}^k \ud\theta \cr
=\sum\limits_{\substack{c_1,\ldots,c_K\\ : u\in c_K}}\xi_{c_1,\ldots,c_K}\frac{1+\sum_{k=1}^K n_{i,j}^{c_k}}{|\set L|+\sum_{k=1}^K n_i^{c_k}}\frac{|\set U|}{n_{c_K}^K}\prod\limits_{k=1..K}\omega_{c_k}^k \label{eq: hat theta K}
\end{eqnarray}
Rearranging \eqref{eq: hat theta K}, we arrive at \eqref{eq: hat theta resem}.

\subsection{Proof of Theorem~\ref{thm: consistency}}
The proof of Theorem~\ref{thm: consistency} relies on the following two lemmas.
\begin{lemma}\label{lem: consistency KL}
If $\mu\in\Pcal(\Theta)$ is in the KL-support of $g$ with respect to $\set H_{\overline{n}}$, then $g(K_{\epsilon,\overline{n}}(\mu)|X^\set U)\underset{|\set U|\to\infty}{\to}1$ for all $\epsilon>0$, $\mu$-almost surely.
\end{lemma}
The above lemma is a perfect analog of a similar statement for Bayesian consistency with direct observations (see \cite{schwartz1965bayes}, Theorem~6.1 and its corollary). The proof also goes through essentially in the same way; therefore, we do not provide it here. This first lemma states that the set $K^C_{\epsilon,\overline{n}}(\mu)$, i.e., the set of distributions $\nu$ that do not agree with the true prior $\mu$ on $\set H_{\overline n}$ according to the KL distance $KL_{\overline n}(\mu,\nu)$ w.r.t. $\set H_{\overline n}$, has a vanishing mass under the posterior distribution $g|X^\set U$, $\mu$-a.s. However, this does not guaranty that the set of distributions $\nu$ with $0<KL_{\overline n}(\mu,\nu)\leq\epsilon$ will have a negligible impact on the estimates $E_g[\theta^u|X^\set U]$. Indeed, for this we need continuity with respect to the KL distance over $\set H_{\overline n}$, which the next lemma provides.
\begin{lemma}\label{lem: guarantee}
Under the assumptions of Lemma~\ref{lem: consistency KL}, for any bounded continuous $f:\Theta\to\mathbb R$,
\begin{equation*}
\lim\limits_{\epsilon\to0}\sup\limits_{\nu\in K_{\epsilon,\overline n}(\mu}\left|E_\nu[f]-\EE[f]\right|=\sup\limits_{\substack{\nu\in\Pcal(\Theta)\\P_\nu=P_\mu\text{ on }\set H_{\overline{n}}}}\left|E_\nu[f]-\EE[f]\right|.
\end{equation*}
\end{lemma}
\bp
Let $\rho$ be the metric on $\Theta$. We use the associated Wasserstein metric $d_\rho$ on $\Pcal(\Theta)$:
\begin{equation*}
d_\rho(\mu,\nu)=\inf\limits_{\substack{\pi\in\Pcal(\Theta^2)\\\pi_1=\mu,\:\pi_2=\nu}}\left\{\int_{(\theta,\lambda)}\rho(\theta,\lambda)\pi(\ud\theta,\ud\lambda)\right\},
\end{equation*}
where $\pi_1$ and $\pi_2$ are the first and second marginals of $\pi$, respectively. It is well-known (see \cite{villani2008optimal}) that the space $\left(\Pcal(\Theta),d_\rho\right)$ is compact, complete and separable, as $(\Theta,\rho)$ is.

Let $\delta>0$ and let $(\epsilon_k)\in\R_+^\N$ be a sequence converging to $0$. For all $k\in\N$, let $\nu_k\in\Pcal(\Theta)$ such that $$\left|E_{\nu_k}[f]-\EE[f]\right|\geq\sup\limits_{\substack{\nu\in\Pcal(\Theta)\\KL_{\overline{n}}(\mu,\nu)\leq\epsilon_k}}\left|E_\nu[f]-\EE[f]\right|-\delta.$$ By compactness of $\left(\Pcal(\Theta),d_\rho\right)$, there exists a converging subsequence $(\widetilde\nu_k)$ of $(\nu_k)$, and a corresponding subsequence $(\widetilde\epsilon_k)$ of $(\epsilon_k)$; let us call $\nu_\infty\in\Pcal(\Theta)$ its limit. Clearly, we have $D_{n^+}(\mu,\nu_\infty)=0$. Because the Wasserstein distance metricizes weak convergence (see Theorem~6.9 in \cite{villani2008optimal}) and $f$ is bounded and continuous, we have that $\lim_{k\to\infty}E_{\widetilde\nu_k}[f]=E_{\nu_\infty}[f]$. Thus,
\begin{multline*}
\sup\limits_{\substack{\nu\in\Pcal(\Theta)\\P_\nu=P_\mu\text{ on }\set H_{\overline{n}}}}\left|E_\nu[f]-\EE[f]\right|\geq\left|E_{\nu_\infty}[f]-\EE[f]\right| \\
=\lim\limits_{k\to\infty}\left|E_{\widetilde\nu_k}[f]-\EE[f]\right|
\geq\lim\limits_{k\to\infty}\sup\limits_{\nu\in K_{\widetilde\epsilon_k,\overline n}(\mu)}\left|E_\nu[f]-\EE[f]\right|-\delta \\
=\lim\limits_{\epsilon\to0}\sup\limits_{\nu\in K_{\epsilon,\overline n}(\mu)}\left|E_\nu[f]-\EE[f]\right|-\delta,
\end{multline*}
where the last inequality is because the sequence is decreasing.
Letting $\delta\to0$ completes the proof.
The opposite inequality is obvious by the definition of $KL_{\overline{n}}(\mu,\nu)$.
\ep

{\noindent \em Proof of Theorem~\ref{thm: consistency}. }
For any bounded continuous $f:\Theta\to\mathbb R$, we have
\begin{multline*}
\Big|E_g[f(\theta^u)|X^\set U]-\EE[f(\theta^u)|X^u]\Big|\leq||f||_\infty g(K^C_{\epsilon,\overline{n}}(\mu)|X^\set U)\\
+\int_{\nu\in K_{\epsilon,\overline{n}}(\mu)}\Big|E_\nu[f(\theta^u)|X^u]-\EE[f(\theta^u)|X^u]\Big|dg(\nu|X^\set U),
\end{multline*}
According to Lemma~\ref{lem: consistency KL}, the first term in the r.h.s. goes to $0$ as $|\set U|\to\infty$, $\mu$-a.s. The second term can always be upper-bounded by $$\sup\limits_{\nu\in K_{\epsilon,\overline n}(\mu)}\Big|E_\nu[f(\theta^u)|X^u]-\EE[f(\theta^u)|X^u]\Big|.$$ By Bayes theorem, $$E_\nu[f(\theta^u)|X^u]=\frac{E_\nu[f(\theta^u)P_{\theta^u}(X^u)]}{P_\nu(X^u)}.$$ For any $x\in\set H_{\overline n}$, Lemma~\ref{lem: guarantee} applied to the bounded continuous function $\theta\mapsto P_{\theta}(x)$ yields $$\lim_{\epsilon\to0}\sup\limits_{\nu\in K_{\epsilon,\overline n}(\mu)}\Big|P_\nu(x)-\PP(x)\Big|=0.$$ Another application of Lemma~\ref{lem: guarantee} to $\theta^u\mapsto \theta^u_{i,j} P_{\theta^u}(X^u)$ completes the proof.
\ep

Note that we could have obtained a version of the Theorem~\ref{thm: consistency} giving a bound on the error in the estimation of any bounded continuous function $f(\theta^u)$ by simply using the function $f(\theta^u)P_{\theta^u}(X^u)$ in the last line of the above proof.

\bibliographystyle{IEEEtran}
\bibliography{IEEEabrv,Reference}

%


\end{document}